\documentclass[10pt,twocolumn,letterpaper]{article}

\usepackage[pagenumbers]{cvpr} %

\usepackage{dsfont}
\usepackage{tikz}
\usetikzlibrary{bayesnet}
\usepackage{color,soul}
\usepackage{multirow}
\usepackage[table]{xcolor}
\usepackage{soul}
\usepackage{minted}
\usepackage{minitoc}
\usepackage[accsupp]{axessibility} 
\usepackage{stfloats} %
\usepackage{tabularx}

\usepackage{tocloft}

\makeatletter
\renewcommand{\paragraph}{%
  \@startsection{paragraph}{4}{\z@}%
    {0.5ex plus 0.2ex minus 0.1ex}%
    {-1em}%
    {\normalfont\normalsize\bfseries}%
}
\makeatother

\newcommand{\highlightrow}{\rowcolor{teal!30}}

\definecolor{cvprblue}{rgb}{0.21,0.49,0.74}
\usepackage[pagebackref,breaklinks,colorlinks,allcolors=cvprblue]{hyperref}

\def\paperID{***} %
\def\confName{CVPR}
\def\confYear{2026}

\title{ViterbiPlanNet: Injecting Procedural Knowledge via \\Differentiable Viterbi for Planning in Instructional Videos}

\author{%
  Luigi Seminara$^{1}$\textsuperscript{$\dagger$} \quad Davide Moltisanti$^{2}$\textsuperscript{*} \quad Antonino Furnari$^{1}$\textsuperscript{*} \\
  $^{1}$University of Catania \quad
  $^{2}$University of Bath \\
  \href{https://gigi-g.github.io/ViterbiPlanNet/}{https://gigi-g.github.io/ViterbiPlanNet/}
}

\begin{document}

\maketitle

\begingroup
\renewcommand{\thefootnote}{\fnsymbol{footnote}}
\setcounter{footnote}{0}

\footnotetext[1]{Equal advising.}
\footnotetext[2]{Work done while visiting the University of Bath.}

\endgroup

\begin{abstract}
Procedural planning aims to predict a sequence of actions that transforms an initial visual state into a desired goal, a fundamental ability for intelligent agents operating in complex environments. Existing approaches typically rely on large-scale models that learn procedural structures implicitly, resulting in limited sample-efficiency and high computational cost. In this work we introduce \textbf{ViterbiPlanNet}, a principled framework that explicitly integrates procedural knowledge into the learning process through a \textbf{Differentiable Viterbi Layer (DVL)}. The DVL embeds a \textbf{Procedural Knowledge Graph (PKG)} directly with the Viterbi decoding algorithm, replacing non-differentiable operations with smooth relaxations that enable end-to-end optimization. This design allows %
the model to learn through graph-based decoding.
Experiments on CrossTask, COIN, and NIV demonstrate that ViterbiPlanNet achieves \textbf{state-of-the-art performance} with an order of magnitude fewer parameters than diffusion- and LLM-based planners. Extensive ablations show that performance gains arise from our differentiable structure-aware training rather than post-hoc refinement, resulting in improved sample efficiency and robustness to shorter unseen horizons. %
We also address testing inconsistencies establishing a unified testing protocol with consistent splits and evaluation metrics. With this new protocol, 
we run experiments multiple times and report results using bootstrapping to assess statistical significance.
\end{abstract}
    
\section{Introduction}
\label{sec:intro}
\begin{figure}[t]
    \centering
    \includegraphics[width=\linewidth]{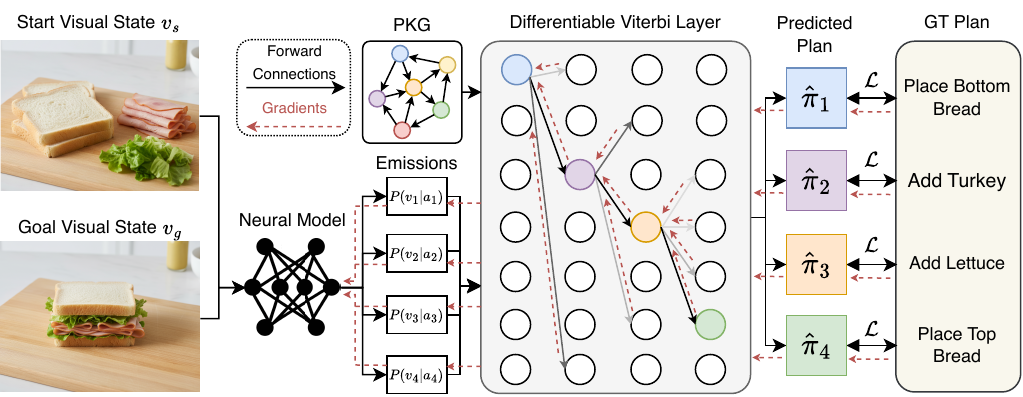}
    \caption{Given start and goal visual states, a {neural model} computes step-wise {emissions}. We propose a {Differentiable Viterbi Layer} that uses a {Procedural Knowledge Graph (PKG)} to decode emissions into a {predicted plan}. The layer allows {gradients} from the planning {Loss ($\mathcal{L}$)} to flow and train the neural model end-to-end, forcing it to learn structure-aware visual representations.}
    \label{fig:teaser}
\end{figure}

Planning a sequence of actions to reach a goal from an initial observation is a fundamental human skill.
For instance, given only the start and goal visual states in Fig.~\ref{fig:teaser}, we can effortlessly infer a likely intermediate plan: \texttt{place bottom bread}, \texttt{add turkey}, \texttt{add lettuce}, and \texttt{place top bread}. 
In this process, we inherently integrate procedural knowledge -- our understanding of valid actions, their preconditions, their effects, and their typical order. This knowledge is what prevents us from planning impossible sequences, such as adding fillings before the first slice of bread is in place.

Replicating planning abilities in artificial systems is crucial for applications like wearable AI assistants that can guide users through complex daily activities. In recent years, this has spurred interest in the task of video procedural planning~\cite{chang2020procedure,wang2023pdpp, niu2024schema, nagasinghe2024not, yang2025planllm,zhou2025masked}: given a start and goal visual observations ($v_s$, $v_g$), predict a plan (a series of actions) to move from $v_s$ to $v_g$.
While researchers have long encoded procedural knowledge explicitly %
in the form of structured graphs~\cite{ashutosh2023video, seminara2024differentiable, seminara2025task, grauman2024ego, lee2024error}, most current planning methods do not~\cite{wang2023pdpp, niu2024schema, yang2025planllm, zhou2025masked}. Instead, methods often resort to implicitly learning complex procedures from large datasets, which is often %
data-inefficient and limits generalization. %
This approach also requires increasingly complex and large models, %
e.g., transformers~\cite{niu2024schema}, LLM-based planners~\cite{yang2025planllm} and diffusion-based sequence generators~\cite{wang2023pdpp, zhou2025masked}. 

We argue that this reliance on implicit learning is a fundamental bottleneck. Instead, we propose to explicitly integrate structured procedural knowledge directly into the planning process during training. %
We encode this knowledge, following previous work~\cite{ashutosh2023video,zhou2023procedure,nagasinghe2024not}, as a Procedural Knowledge Graph (PKG): a directed graph where nodes are actions (e.g., \texttt{place bottom bread}), edges are transitions, and edge weights denote transition probabilities (e.g., \texttt{place bottom bread} $\overset{0.8}{\to}$ \texttt{add turkey}).
This allows us to reframe procedure planning as the problem of decoding the most likely sequence of hidden states (actions) that explain a sequence of observed events (the start and goal visual input).
This problem is typically addressed with the Viterbi algorithm~\cite{1054010}. However, previous work has used Viterbi merely as a post-processing technique~\cite{niu2024schema, yang2025planllm, zhao2022p3iv}, which %
treats the graph as a correcting mechanism, rather than a guide. 
We hypothesize that by fully integrating the Viterbi algorithm in the training process we can relieve the model from having to \textit{extract and memorize domain-specific procedural knowledge}, enabling the design of simpler parameter-efficient approaches. 
To validate this we introduce ViterbiPlanNet, a novel framework that embeds the PKG directly into the planning algorithm. This end-to-end integration is enabled by the introduction of a Differentiable Viterbi Layer (DVL), which replaces non-differentiable \texttt{max} and \texttt{argmax} operations with differentiable relaxations~\cite{mensch2018differentiable}. 
Integrating this layer allows gradients to flow from the predicted plan back to the neural model. 
This fundamentally simplifies the model's task: instead of being forced to learn and predict the entire complex plan, the model is now only responsible for predicting emission probabilities—i.e., the compatibility of a given action with the visual observations. At inference, this approach %
guarantees a plan that is structurally consistent with the learned procedural graph.

We evaluate our approach on three standard datasets: CrossTask~\cite{zhukov2019cross}, COIN~\cite{tang2019coin}, and NIV~\cite{alayrac2016unsupervised}.
Recent work~\cite{niu2024schema,zhou2025masked} highlighted inconsistencies in training and evaluation settings in the literature~\cite{wang2023pdpp, niu2024schema, nagasinghe2024not, yang2025planllm}, however to date these inconsistencies remain unaddressed. We thus establish and open-source a unified evaluation pipeline and re-benchmark prior methods under consistent conditions, averaging performance over multiple runs and reporting confidence intervals for performance estimates and performance differences.
Results on this unified benchmark validate our approach. We show that ViterbiPlanNet, despite its simpler and parameter-efficient design, consistently and significantly outperforms all re-benchmarked prior methods. %
Our ablations confirm that the differentiable end-to-end training is critical, yielding far greater gains than using Viterbi as a post-processing decoder or for conditioning a diffusion model. Our approach makes ViterbiPlanNet highly sample-efficient %
and able to make predictions at planning horizons shorter than those seen during training.
In sum, our contributions are: %
\begin{itemize}
    \item We introduce \textit{ViterbiPlanNet}, a novel framework that integrates a Procedural Knowledge Graph (PKG) end-to-end via a \textit{Differentiable Viterbi Layer}. This design is inherently lightweight, enabling our model to learn simple emission probabilities in a parameter and sample-efficient way, rather than memorizing complex procedural rules.
    \item We establish and open-source a \textit{standardized evaluation benchmark}, which unifies data splits and evaluation metrics implementations, providing a fair and rigorous comparison of state-of-the-art methods, addressing key inconsistencies in prior work.
    \item We introduce a novel \textit{cross-horizon} testing protocol to check for consistency where models are tested on shorter horizons than the ones they were trained on.
\end{itemize}

\section{Related Work}
\label{sec:related_work}

\paragraph{Procedure Planning in Instructional Videos.}
Early procedure planning models required full supervision on intermediate visual steps~\cite{chang2020procedure, sun2022plate, bi2021procedure}, 
while more recent work leveraged language or high-level structure instead of dense frame annotations~\cite{zhao2022p3iv, li2023skip, wang2023event}.
Recent state-of-the-art methods proposed architectures based on diffusion models~\cite{wang2023pdpp, nagasinghe2024not, zhou2025masked}, Large Language Models (LLMs)~\cite{yang2025planllm} or language-aligned transformer-based architectures~\cite{niu2024schema}. 
These models encode procedural knowledge implicitly in their parameters rather than in explicit external structures.

Graph-based reasoning has long been central to classical planning~\cite{bryce2007tutorial}, yet integrating such structure in modern procedure planning algorithms is under-investigated. Few previous methods use procedural graphs as retrieval signals~\cite{nagasinghe2024not} or for Viterbi Decoding post-processing at inference-time~\cite{niu2024schema, yang2025planllm}. In contrast, our approach directly integrates procedural knowledge via a novel \emph{Differentiable Viterbi Layer}, enabling end-to-end learning of emission probabilities directly from visual observations, which are used to predict a procedural sequence by referencing a Procedural Knowledge Graph (PKG). This yields a lightweight yet structured planner enforcing global coherence without relying on massive diffusion or LLM-based architectures.

\paragraph{Evaluation of Procedural Planning Approaches.} 
Previous work~\cite{niu2024schema, zhou2025masked} noted important evaluation inconsistencies, including different training and testing protocols~\cite{nagasinghe2024not}, inconsistent implementations of evaluation metrics~\cite{niu2024schema}, different feature extraction schemes~\cite{wang2023pdpp} and data loaders~\cite{niu2024schema, wang2023pdpp, nagasinghe2024not, zhou2025masked}, as well as vastly different parameter counts~\cite{yang2025planllm, zhou2025masked}. 
We also found that some methods exhibit large performance variations when trained with different seeds. These factors hinder fair comparison. %

To address this issue, we propose and open-source a unified evaluation protocol where we run experiments multiple times and report confidence intervals to assess statistical significance. We anticipate that this unified protocol will support future research, enable fair comparison, and provide a more rigorous progress assessment.

\paragraph{Explicit Procedural Knowledge in Computer Vision.}
Understanding complex goal-directed activities depends on the ability of models to capture the underlying procedural structure of a task. A common representation for such structure is the task graph, a human-interpretable graph where nodes denote key procedural steps and edges encode their temporal or causal dependencies~\cite{seminara2024differentiable, grauman2024ego, ashutosh2023video, seminara2025task}.
Early work constructed task graphs from textual instructions, such as recipes, using rule-based method~\cite{zhou2023procedure}. Later studies proposed to infer graphs directly from video data, leveraging co-occurrence statistics to capture the relationships among procedural steps~\cite{ashutosh2023video}. Recent studies moved toward learning these structures from video inputs in a differentiable fashion~\cite{seminara2024differentiable, seminara2025task}.
Explicit task graphs have also become pivotal in enabling structured reasoning across a variety of downstream tasks. In the Ego-Exo4D benchmark~\cite{grauman2024ego}, manually annotated graphs %
provide the foundation for evaluating higher-level reasoning tasks such as missing-step prediction, procedural mistake detection, and next-step anticipation. Beyond benchmarking, incorporating graph-based priors has proven effective in online action segmentation%
~\cite{shen2024progress} and error detection%
~\cite{lee2024error, seminara2025task, seminara2024differentiable, lee2025error}. %

While previous work has mainly used graphs for procedure planning as priors~\cite{nagasinghe2024not} or for post-processing~\cite{zhao2022p3iv, niu2024schema, yang2025planllm}, we integrate procedural guidance directly into the learning process through a Differentiable Viterbi Layer, enforcing graph-consistent reasoning at training time.

\paragraph{Viterbi for Procedure Planning.} 
The Viterbi algorithm~\cite{1054010} is a classical dynamic programming method used to recover the most probable sequence of latent states given a series of observations. Previous procedure planning work~\cite{zhao2022p3iv, niu2024schema, yang2025planllm} adopted it as a post-processing step to refine action predictions at inference time. While this post-processing improves temporal consistency, it does not leverage the structural priors encoded in the Viterbi logic for training.
Building on differentiable dynamic programming~\cite{mensch2018differentiable}, we propose a \textit{Differentiable Viterbi Layer} (DVL) that replaces the non-differentiable \textit{max} and \textit{argmax} operators with smooth relaxations. This enables us to integrate a differentiable decoding routine into our planning model, guiding the learning of a neural network that predicts emission probabilities from visual inputs. 

Unlike~\cite{mensch2018differentiable} which jointly learns both transition and emission parameters, our DVL introduces no additional trainable parameters, leveraging 
\emph{pre-computed procedural knowledge} in the form of a fixed transition matrix estimated from action co-occurrence statistics~\cite{ashutosh2023video}. Our DVL %
outputs a soft plan that is composed recursively from the computed \emph{soft backpointer distribution}. %

\begin{figure*}[t]
    \centering
    \includegraphics[width=0.99\linewidth]{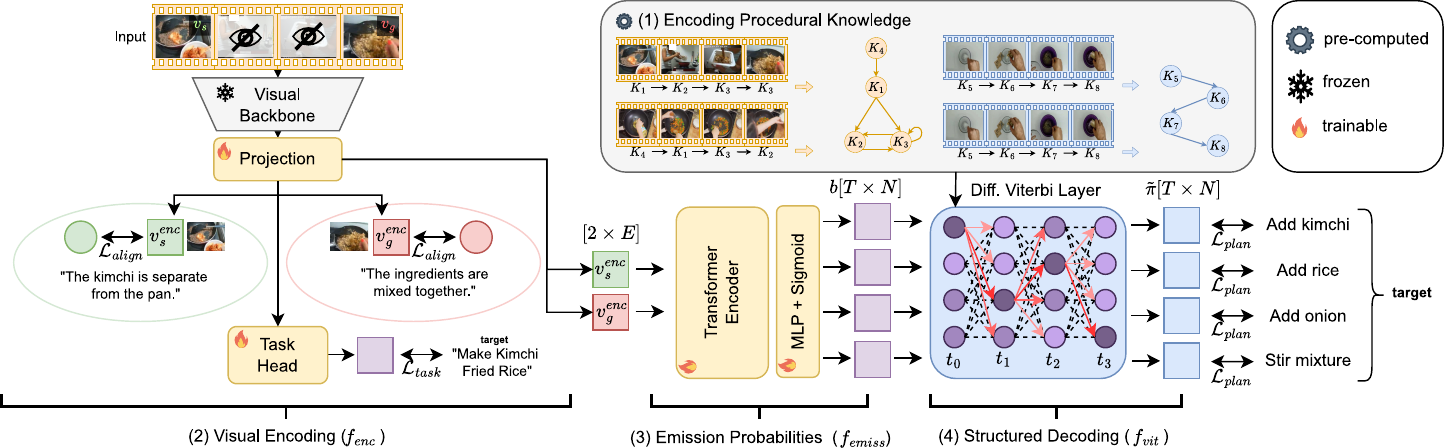}
    \caption{ViterbiPlanNet consists of four main stages: 1) Encoding Procedural Knowledge -- extracting PKGs from training data 2) Visual Encoding ($f_{enc}$) -- extracting features from start and goal frames, trained with the $\mathcal{L}_{align}$ and $\mathcal{L}_{task}$ losses, 3) Computing emission probabilities $b$ with $f_{emiss}$, and 4) Structured Decoding ($f_{vit}$) -- parametrized by the PKG, taking as input emission probabilities and outputting a soft plan $\tilde{\pi}$. Training with a plan loss $\mathcal{L}_{plan}$, gradients pass through Structured Decoding and optimize $f_{emiss}$. 
    }
    \label{fig:architecture}
\end{figure*}

\section{Method}
\label{sec:method}

\paragraph{Problem Formulation.}
Let $\mathcal{K} = \{K_1, K_2, \dots, K_{N}\}$ be the action taxonomy.
Given visual start state $v_s$, visual goal state $v_g$, and a target plan length $T$, the objective is to generate an optimal plan $\pi^* \in \mathcal{K}^T$, i.e., a sequence of $T$ actions $(a_1, a_2, \dots, a_T)$ allowing to reach $v_g$ starting from $v_s$.
We assume a probabilistic graphical model where latent actions $(a_0,a_1,\ldots, a_T)$ generate visual states $(v_s=v_0, v_1, \ldots, v_{T-1},v_{T}=v_g)$ and influence subsequent actions (i.e., $a_t \to a_{t+1}$). Among visual states, the start and goal states ($v_0$ and $v_T$) are observed, while others are latent.

\resizebox{0.91\hsize}{!}{
\begin{tikzpicture}
  \node[latent] (a0) at (0, 0) {$a_0$};
  \node[latent] (a1) at (2, 0) {$a_1$};
  \node (dots) at (4, 0) {$\dots$};
  \node[latent] (aT-1) at (6, 0) {$a_{T-1}$};
  \node[latent] (aT) at (8, 0) {$a_T$};

  \node[obs] (vs) at (0, -1.5) {$v_0$};   %
  \node[latent] (v1) at (2, -1.5) {$v_1$};
  \node[latent] (vT-1) at (6, -1.5) {$v_{T-1}$};
  \node[obs] (vT) at (8, -1.5) {$v_T$};  %

  \edge {a0} {a1};
  \edge {a1} {dots};
  \edge {dots} {aT-1};
  \edge {aT-1} {aT};

  \edge {a0} {vs};
  \edge {a1} {v1};
  \edge {aT-1} {vT-1};
  \edge {aT} {vT};
\end{tikzpicture}
}
Assuming %
that an action $a_t$ is dependent only on the past action $a_{t-1}$ (the Markov property\footnote{\label{fn:supp}See the supplementary material for more details.}),
we can write the joint probability as follows:
\begin{equation}
\small
P(a_{0:T}, v_{0:T})  = P(a_0) P(v_0 | a_0) \prod_{t=1}^{T} P(a_t | a_{t-1}) P(v_t| a_t).
\end{equation}
Since we are not interested in estimating $a_0$ (the action leading to $v_s$) and considering that visual states $v_{0:T}$ are fixed at inference, we can rewrite the expression above as follows$^{\ref{fn:supp}}$:
\begin{equation}
\small
    P(\pi = a_{1:T}| v_{0:T}) \propto  \prod_{t=1}^T {P(a_t | a_{t-1})} {P(v_{t}| a_t)}.
    \label{eq:conditional}
\end{equation}
We seek the plan $\pi^*$ maximizing Eq.~\eqref{eq:conditional}:
\begin{equation}
    \pi^*
    = \underset{\pi =a_{1:T} \in \mathcal{K}^T}{\arg\max} \; \prod_{t=1}^T \underbrace{P(a_t | a_{t-1})}_{\text{Transition}} \underbrace{P(v_{t}| a_t)}_{\text{Emission}}.
    \label{eq:maximization}
\end{equation}
Notably, this maximization problem can be solved using the Viterbi algorithm~\cite{1054010}, as discussed in the following.

\paragraph{ViterbiPlanNet.}
We define ViterbiPlanNet (see Fig.~\ref{fig:architecture}) based on the probabilistic framework defined in Eq.~\eqref{eq:maximization}, which allows us to decompose the prediction problem into four stages: 1) \textbf{Encoding Procedural Knowledge}, 2) \textbf{Visual Encoding}, 3) \textbf{Emission Probabilities}, 4) \textbf{Structured Decoding}. These stages are presented in the following and more details are given in the supplementary material.

\paragraph{Encoding Procedural Knowledge.}
We encode a procedure %
with a pre-computed \textit{Procedural Knowledge Graph} (PKG)  $\mathcal{G} = (\mathcal{V}, \mathcal{E}, \omega)$, where nodes are actions, i.e., $\mathcal{V} = \mathcal{K}$, directed edges $\mathcal{E} \subseteq \mathcal{V} \times \mathcal{V}$ represent valid transitions, and function $\omega:\mathcal{E} \rightarrow [0,1]$ assigns each edge a transition probability, which we estimate based on the co-occurrence of actions in the training set as in~\cite{ashutosh2023video}.
We model transition probabilities in Eq.~\eqref{eq:maximization} based on the graph: $P(a_t|a_{t-1}) = \omega(a_{t-1},a_t)$\footnote{See the supplementary material for dependence on PKG quality.}. We compute a different graph per dataset\footnote{See the supplementary material for results using a single PKG.}.

\paragraph{Visual Encoding.} Our input is a pair of short video clips $v_s$ and $v_g$ capturing the start and goal visual states\footnote{See the supplementary material for experiments using intermediate visual observations.}.
We encode these %
states with a visual encoding function $f_{enc}$ following~\cite{niu2024schema}:
\begin{align}
    v_s^{enc} = f_{enc}(v_s) \in \mathbb{R}^E \qquad
    v_g^{enc} = f_{enc}(v_g) \in \mathbb{R}^E.
\end{align}
This involves extracting features with a frozen visual backbone, followed by a learnable projection\footnote{The supplementary material reports a more detailed description.\label{fn:supp3}}. %

\paragraph{Emission Probabilities.} Emission probabilities in Eq.~\eqref{eq:maximization} depend on visual states $v_t$, which, except for the start and goal states, are unobserved. Hence, we propose to predict $P(v_{t}| a_t)$ from the start/goal states with a network $f_{emiss}$:
\begin{equation}
    P(v_{t}| a_t=K_j) = f_{emiss} (v^{enc}_s,v^{enc}_g; t, j).
\end{equation}
In practice, we design $f_{emiss}$ as a transformer encoder followed by an MLP and a sigmoid activation, predicting a matrix $b \in T \times N$, i.e., $b=f_{emiss}(v^{enc}_s,v^{enc}_g)$. %

\paragraph{Structured Decoding.}
The maximization problem in Eq.~\eqref{eq:maximization} is notably solved with the Viterbi algorithm~\cite{1054010}, which decodes the optimal sequence of actions based on the transition and emission probabilities. Standard Viterbi decoding is non-differentiable and prevents end-to-end training, including the tuning of the $f_{emiss}$ network. Hence, in the following section we propose a Differentiable Viterbi Layer which we denote as $f_{vit}$. %

\paragraph{Differentiable Viterbi Layer (DVL).}
Standard Viterbi decoding~\cite{1054010} relies on standard $\max$ and $\arg\max$ operations, which make it non-differentiable and therefore unsuitable for end-to-end training. To overcome this limitation, we adopt the log-sum-exp and softmax relaxations introduced in~\cite{mensch2018differentiable}. Given a vector $\mathbf{x}\in \mathbb{R}^N$, $\mathrm{S\text{-}max}(\mathbf{x})\in \mathbb{R}$ returns a differentiable estimate of the maximum value of $\mathbf{x}$, and $\mathrm{S\text{-}argmax}(\mathbf{x}) \in [0,1]^N$ returns a probability distribution over indices that reflects their relative proximity to the maximum$^{\ref{fn:supp3}}$.
Following Viterbi decoding~\cite{1054010}, at each time step $t$ we define state scores $\delta_t(j)$ representing the cumulative score of reaching state $j$ at time $t$. At $t=1$ there are no previous actions to condition on, so we initialize the state scores directly from the emission probabilities: $\delta_1(j) = b[1,j]$. For subsequent steps ($t>1$) we compute predecessor scores $s_{i\to j}^{(t)} = \delta_{t-1}(i)\,\omega(i,j)$ for each possible transition from state $i$ to state $j$, where $\omega(i,j) = P(a_j|a_i)$ are our fixed transition probabilities derived from the PKG. State scores are then updated by applying the smooth max operator to the set of predecessor scores, modulated by the emission score $b[t,j]$:
\begin{equation}
    \delta_t(j) = b[t,j]\;\mathrm{S\text{-}max}\big(\{s_{i\to j}^{(t)}\}_{i=1}^N\big).
    \label{eq:emissions}
\end{equation}
In parallel, we compute a \textit{soft backpointer distribution} $\boldsymbol{\psi}_t(j,\cdot)\in[0,1]^N$ over predecessor actions, serving as the differentiable equivalent of a set of discrete backpointers:
\begin{equation}
    \boldsymbol{\psi}_t(j,k) = \mathrm{S\text{-}argmax}\big(\{s_{i\to j}^{(t)}\}_{i=1}^N\big)_k, k = 1, \dots, N. 
    \label{eq:soft_backpointers}
\end{equation}
During the backward pass these soft backpointers are recursively composed to produce a \emph{soft plan} $\tilde\pi \in[0,1]^{T\times N}$, i.e., a time-indexed distribution over actions that smoothly approximates the discrete Viterbi solution$^{\ref{fn:supp3}}$.
It is worth noting that the proposed Differentiable Viterbi Layer (DVL) does not introduce additional training parameters: it is parametrized by fixed transition probabilities $\omega(i,j)$ (the PKG), takes as input emission probabilities $b$ (computed with $f_{emiss}$), and outputs the \textit{soft plan} $\tilde \pi \in[0,1]^{T\times N}$.

\paragraph{Training.} The main optimization signal is the \emph{planning loss} $\mathcal{L}_{plan}$, which directly supervises the output of the DVL minimizing the Mean Squared Error (MSE) between the predicted soft plan $\tilde \pi \in [0,1]^{T \times N}$ and the ground-truth one-hot plan $\tilde \pi^{GT}$. This term ensures that the Differentiable Viterbi Layer (DVL) learns to generate graph-consistent action sequences that match the target plans. Two additional standard losses provide auxiliary supervision for the visual encoder $f_{enc}$: (1) the \emph{visual-semantic alignment loss}~\cite{niu2024schema} $\mathcal{L}_{align}$, which encourages alignment between visual embeddings and textual descriptions of procedural states, and
(2) the \emph{task classification loss}~\cite{wang2023pdpp} $\mathcal{L}_{task}$, which guides the encoder to preserve global task-level semantics by predicting the procedure label from visual observations\footnote{More details in the supplementary material.\label{fn:supp6}}.
The final objective combines these three terms with equal weights:
\begin{equation}
\mathcal{L} = \mathcal{L}_{plan} + \mathcal{L}_{align} + \mathcal{L}_{task}.
\label{eq:total_loss}
\end{equation}

\paragraph{Inference.} At inference the model takes as input the initial state $v_s$, the goal state $v_g$, and the procedural knowledge graph $\mathcal{G}$ (the same used during training), and returns the soft plan $\tilde \pi \in [0,1]^T$. 
We hence use the standard Viterbi Decoding (VD) to derive a discrete plan $\pi \in \mathcal{K}^N$ as done in prior work~\cite{niu2024schema, zhao2022p3iv, yang2025planllm}, unless otherwise stated.
This choice ensures that the predicted plan corresponds to the most probable path consistent with both visual evidence and structural constraints encoded in $\mathcal{G}$.

\section{Experiments and Results}
\paragraph{Datasets and Metrics.} We evaluate our proposed ViterbiPlanNet on three standard benchmarks for procedure planning\footnote{We report results on EgoPER~\cite{lee2024error} in the supplementary material.}: \textbf{CrossTask}~\cite{zhukov2019cross}, which provides 2,750 videos for 18 tasks; \textbf{COIN}~\cite{tang2019coin}, a large-scale dataset with 11,827 videos covering 180 tasks; and \textbf{NIV}~\cite{alayrac2016unsupervised}, a smaller dataset of 150 videos for 5 tasks. 
We test for $T \in [3,4]$, depending on the experiments\footnote{We report results with $T\in[5, 6]$ in the supplementary material.}.
We adopt standard evaluation metrics~\cite{bi2021procedure, niu2024schema}: (1) the \textit{Success Rate} (SR) measures the percentage of predicted sequences that match the ground-truth sequence exactly;  
(2) the \textit{Mean Accuracy} (mAcc) computes the average step-wise accuracy, i.e., the proportion of correctly predicted actions at each time step; and  
(3) the \textit{Mean Intersection over Union} (mIoU) quantifies the overlap between predicted and ground-truth action sequences.

\paragraph{Unified Evaluation Protocol.}
We propose a unified evaluation protocol where all methods are re-trained and evaluated to make use of standardized experimental settings\footnote{We do not re-train LLM-based models and MTID~\cite{zhou2025masked} because they exceed our computational budget.}. We use or adapt the official implementation of all methods to run experiments.
To assess statistical significance of performance improvements, we train each model with five random seeds and report mean performance and performance differences with 90\% confidence intervals, computed using bootstrapping (denoted as $xx \pm yy$). We hence mark as statistically significant only performance differences whose confidence interval does not include zero$^{\ref{fn:supp6}}$. Parameter counts or estimates thereof are reported to contextualize performance with model capacity. We believe this unified protocol provides a significant contribution %
to support research on this task$^{\ref{fn:supp6}}$.

\begin{table}[t]
\centering
\caption{Ablation of Viterbi components on CrossTask for $T{=}3$.}
\label{tab:ablation_reorg}
\resizebox{0.97\linewidth}{!}{%

\begin{tabular}{
    c
    >{\centering\arraybackslash}p{0.9cm}  %
    >{\centering\arraybackslash}p{0.9cm}  %
    >{\centering\arraybackslash}p{0.9cm}  %
    r r r
}

\toprule
 & \multicolumn{1}{c}{\textbf{Train}} & \multicolumn{2}{c}{\textbf{Inference}} & \multicolumn{3}{c}{\textbf{Metrics (\%) ↑}} \\
\cmidrule(lr){2-2} \cmidrule(lr){3-4} \cmidrule(l){5-7}
 & \textbf{DVL} & \textbf{DVL} & \textbf{VD} & \multicolumn{1}{c}{\textbf{SR}} & \multicolumn{1}{c}{\textbf{mAcc}} & \multicolumn{1}{c}{\textbf{mIoU}} \\
\midrule
1 & $\times$ & $\times$ & $\times$ & 32.47  \scriptsize $\pm$ 0.32 & 60.63  \scriptsize $\pm$ 0.22 & 82.45  \scriptsize $\pm$ 0.02 \\
2 & $\times$ & $\times$ & \checkmark & 32.99  \scriptsize $\pm$ 0.28 & 58.57  \scriptsize $\pm$ 0.20 & 82.34 \scriptsize $\pm$ 0.08 \\
3 & $\times$ & \checkmark & $\times$ & 32.09  \scriptsize $\pm$ 0.26 & 58.57  \scriptsize $\pm$ 0.20 & 82.34  \scriptsize $\pm$ 0.08 \\
4 & $\times$ & \checkmark & \checkmark & 30.77  \scriptsize $\pm$ 0.19 & 57.04  \scriptsize $\pm$ 0.12 & 81.95  \scriptsize $\pm$ 0.20 \\
\cmidrule(lr){1-7}
5 & \checkmark & $\times$ & $\times$ & 20.05  \scriptsize $\pm$ 0.63 & 54.61  \scriptsize $\pm$ 0.27 & 76.99  \scriptsize $\pm$ 0.41 \\
6 & \checkmark & $\times$ & \checkmark & \underline{38.09}  \scriptsize $\pm$ 0.39 & 63.05  \scriptsize $\pm$ 0.30 & \underline{83.83}  \scriptsize $\pm$ 0.22 \\
7 & \checkmark & \checkmark & $\times$ & 37.66  \scriptsize $\pm$ 0.45 & \textbf{63.15}  \scriptsize $\pm$ 0.21 & 83.81  \scriptsize $\pm$ 0.12 \\
8 & \checkmark & \checkmark & \checkmark & \textbf{38.45}  \scriptsize $\pm$ 0.32 & \underline{63.07}  \scriptsize $\pm$ 0.17 & \textbf{83.89}  \scriptsize $\pm$ 0.16 \\
\midrule
\multicolumn{4}{l}{Improvement w.r.t. conf. $1$} & 5.98 \scriptsize $\pm$ 0.47 & 2.44 \scriptsize $\pm$ 0.29 & 1.44 \scriptsize $\pm$ 0.16 \\
\bottomrule
\end{tabular}%
}
\end{table}

\subsection{Ablations on CrossTask ($T=3$)}
In this section, we present ablation studies and analysis to demonstrate the advantages of incorporating structure-aware training through the proposed probabilistic framework. We report results on CrossTask with $T{=}3$, and provide additional ablations in the supplementary material.

\paragraph{Importance of Structure-Aware Training.} Table~\ref{tab:ablation_reorg} compares different configurations of ViterbiPlanNet on CrossTask for $T{=}3$, where the Differentiable Viterbi Layer (DVL) and standard Viterbi Decoding (VD) are included or excluded at training or inference. %
Configurations 5-8 use the proposed DVL at training time, but perform inference in different ways: removing the DVL and taking the argmax directly on emission probabilities (5), performing standard VD on emission probabilities (6), taking the argmax on the soft plan produced by the DVL (7), and applying VD on top of the soft plan produced by the DVL (8).
Configurations 1-4 denote a base model trained to predict the soft plan directly from the emission probabilities without the DVL at training time. Inference is performed: with an argmax on the predicted soft plan (1), decoding the soft plan with standard VD (2), post-processing the soft plan with the DVL and taking the argmax (3), and applying VD on top of the soft plan refined by the DVL (4).
Results highlight the following findings. \\
\ul{Structure-aware training is effective}. Our full models ($6-8$) obtain absolute improvements of $\approx 6\%$ in \textbf{SR} with respect to the baseline ($1$). Adding standard VD ($2$), DVL ($3$) or both ($4$) to the base model at inference does not bring substantial improvements, showing that DVL guides training, rather than serving merely as a post-processing step.
\\
\ul{DVL Learns Meaningful Emissions.}
Using emissions learned with DVL
directly for inference %
leads to poor results ($5$). %
Indeed, emissions %
are distributions over states, not actions, hence they are unsuitable for direct predictions, but lead to good results with DVL ($7$) or standard VD ($6$). We expand on this in Fig.~\ref{fig:qualitative_result} with a qualitative analysis.\\
\ul{DVL is Backward-Compatible with Standard VD}. Replacing DVL with VD at inference does not lead to substantial performance differences, either when DVL is used for training ($6$ vs $7$), or not ($3$ vs $2$).
Adding VD on top of DVL brings comparable performance ($7$ vs $8$).

\paragraph{Memorization and Sample Efficiency.}
We postulate that part of the advantage of using complex architectures~\cite{niu2024schema} for procedure planning consists in their ability to memorize procedural knowledge. In contrast, rather than memorizing an entire procedure, we learn to predict the optimal plan path \textit{step-by-step} guided by the PKG.
This brings benefits in terms of sample efficiency (i.e., fewer example sequences are needed at training time).
To assess this, we compare ViterbiPlanNet to SCHEMA~\cite{niu2024schema}, a model with a comparable parameter count ($\approx 6M$) which is built on a more complex transformers-based architecture. 
To isolate the contribution of the planning module, both methods use the same visual encoder $f_{enc}$ pre-trained on the whole training set\footnote{Using $f_{enc}$ within SCHEMA requires no modifications.} and the same PKG. We then freeze the visual encoder and train the planning component using progressively larger fractions of the training data. We compare models when using the PKG (ViterbiPlanNet vs SCHEMA, which by default uses Viterbi Decoding for post-processing) and when not (Base Model vs SCHEMA w/o Viterbi Decoding). The Base Model here corresponds to conf. 1 in Table~\ref{tab:ablation_reorg}.

\begin{figure}[t]
    \centering
    \includegraphics[width=1.0\linewidth]{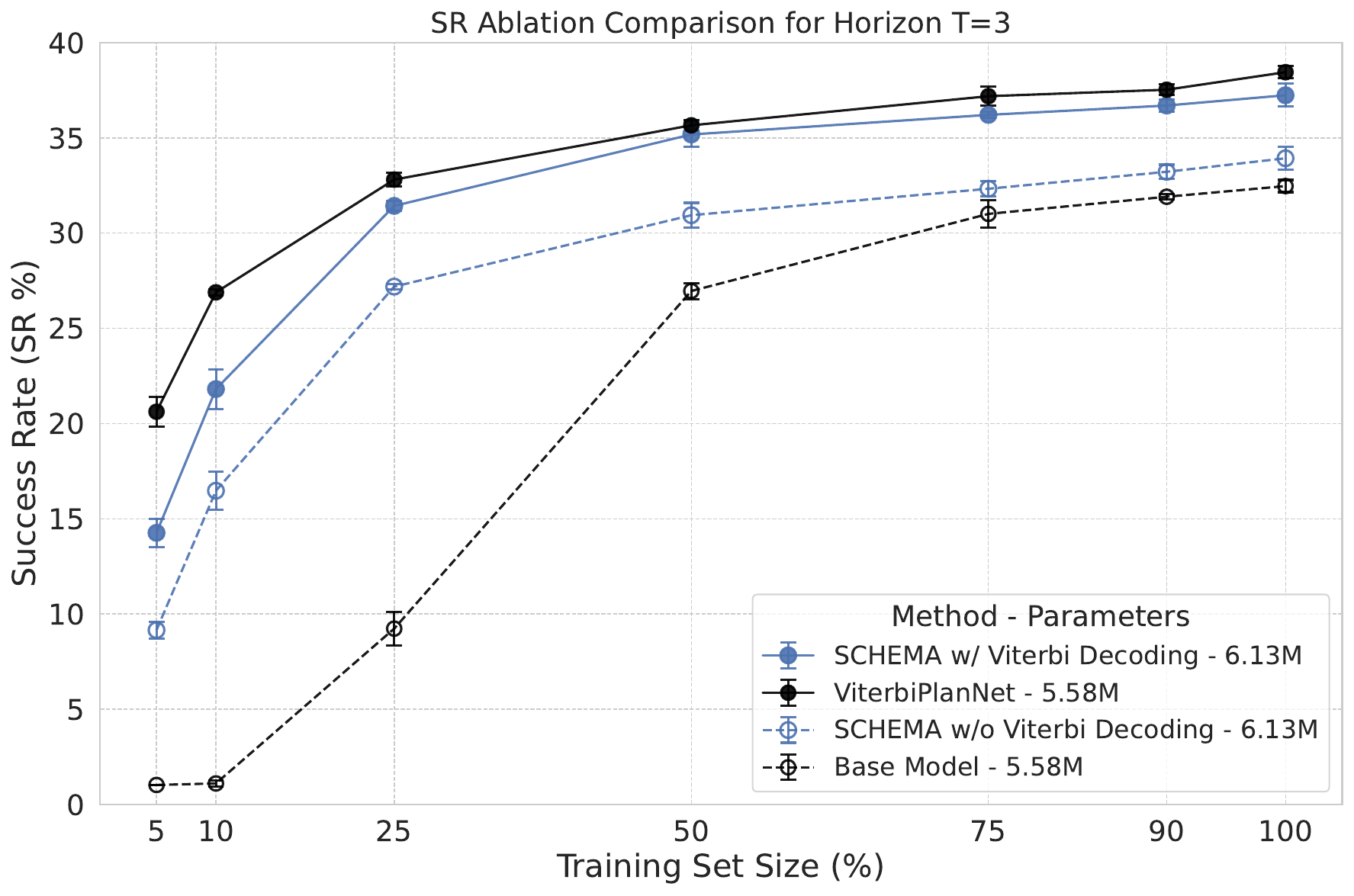}
    \caption{Performance as a function of training data on CrossTask for $T=3$. \textit{ViterbiPlanNet} is more parameter-efficient, as it does not need to memorize procedural knowledge.
    }
    \label{fig:memorization}
\end{figure}

Results in Fig.~\ref{fig:memorization} show that %
ViterbiPlanNet is more sample-efficient, achieving better results with fewer examples. The gap between ViterbiPlanNet and SCHEMA decreases with more training examples, which favors memorization.
When the PKG is removed (dashed lines), SCHEMA achieves better performance compared to the base model due to its more flexible architecture, allowing for better memorization of procedural knowledge. 
Importantly, the Base Model and ViterbiPlanNet have the same parameter count and architecture, hence the improvement of ViterbiPlanNet is entirely due to its PKG-awareness during training%
, which does not require memorization. %

\paragraph{Guided Training vs Conditioning and Post-processing.} We compare the way we use the PKG at training time (\textit{guided training}) against other prior approaches that leverage a PKG in procedure planning: 
KEPP~\cite{nagasinghe2024not} uses PKGs to sample candidate paths that \textit{condition} a diffusion model, while SCHEMA~\cite{niu2024schema} and PlanLLM~\cite{yang2025planllm} apply the PKG as a \textit{post-processing} step with classical Viterbi Decoding (VD)~\cite{1054010}. 
Table~\ref{tab:sr_comparison_pkg} compares Success Rate 
when using or bypassing the PKG\footnote{See more details in the supplementary material.}.
For our method, the ``w/o PKG'' configuration coincides with the Base Model (conf. 1 in Table~\ref{tab:ablation_reorg}). %
Results highlight two key observations. First, \ul{all methods benefit from the PKG}, confirming its value as a source of procedural structure. Second, \ul{ViterbiPlanNet benefits the most}. Specifically, our model improves by +5.98\% SR when the PKG is enabled, while SCHEMA (+3.31\%), PlanLLM (+1.97\%), and KEPP (+2.56\%) exhibit smaller gains. This suggests that learning through the PKG via a Differentiable Viterbi Layer (DVL) is significantly more effective than relying on structural constraints only at inference time.

\begin{table}[t]
\centering
\caption{SR $\uparrow$ (\%) with and without PKG on CrossTask for $T=3$.}
\label{tab:sr_comparison_pkg}
\resizebox{\linewidth}{!}{%
\begin{tabular}{l c c c c}
\toprule
\textbf{Method} & \textbf{PKG Use} & \textbf{w/o PKG} & \textbf{w/ PKG} & \textbf{Improv.} \\
\midrule
KEPP~\cite{nagasinghe2024not} & Conditioning & 32.37 \scriptsize $\pm$ 5.36 & 34.93 \scriptsize $\pm$ 2.60 & 2.56 \scriptsize $\pm$ 5.93\\
PlanLLM~\cite{yang2025planllm} & Post-processing & 34.87 \scriptsize $\pm$ 1.32 & 36.84 \scriptsize $\pm$ 1.21 & 1.97 \scriptsize $\pm$ 1.81\\
SCHEMA~\cite{niu2024schema} & Post-processing & 33.93 \scriptsize $\pm$ 0.59 & 37.24 \scriptsize $\pm$ 0.60 & 3.31 \scriptsize $\pm$ 0.84 \\
\textbf{ViterbiPlanNet} & Guided Training & 32.47 \scriptsize $\pm$ 0.32 & 38.45 \scriptsize $\pm$ 0.32 & 5.98 \scriptsize $\pm$ 0.47 \\
\bottomrule
\end{tabular}%
}
\end{table}

\begin{table*}[t]
\centering
\caption{Comparison with the state of the art. \textbf{Best} and \underline{second-best} results are highlighted for each metric within each time horizon. Statistically significant performance differences (i.e., cases in which the confidence interval does not include zero) are {\setlength{\fboxsep}{0pt}\colorbox{yellow!30}{marked in yellow}}.}
\label{tab:performance_combined_viterbiplannet}
\resizebox{\textwidth}{!}{%
\begin{tabular}{@{}ll lllr lllr lllr@{}}
\toprule
\multirow{2}{*}{\textbf{T}} & \multirow{2}{*}{\textbf{Method}} & \multicolumn{4}{c}{\textbf{CrossTask}} & \multicolumn{4}{c}{\textbf{COIN}} & \multicolumn{4}{c}{\textbf{NIV}} \\
\cmidrule(lr){3-6} \cmidrule(lr){7-10} \cmidrule(lr){11-14}
& & SR $\uparrow$ (\%) & mAcc $\uparrow$ (\%) & mIoU $\uparrow$ (\%) & Params (M) & SR $\uparrow$ (\%) & mAcc $\uparrow$ (\%) & mIoU $\uparrow$ (\%) & Params (M) & SR $\uparrow$ (\%) & mAcc $\uparrow$ (\%) & mIoU $\uparrow$ (\%) & Params (M) \\
\midrule
\multirow{11}{*}{3} 
& Qwen2.5-VL-32B~\cite{bai2025qwen2} & 11.48 & 36.35 & 69.52 & 32,000 & 3.65 & 17.51 & 52.10 & 32,000 & 7.41 & 27.65 & 59.73 & 32,000 \\
& Qwen2.5-32B~\cite{qwen2024qwen2} & 25.14 & 56.10 & 80.92 & 32,000 & 14.97 & 36.34 & 78.74 & 32,000 & 24.07 & 43.46 & 71.88 & 32,000 \\
& Gemini 2.5 Pro~\cite{comanici2025gemini} & 29.18 & 57.90 & 81.48 & $>$100,000 & 17.02 & 38.87 & 78.73 & $>$100,000 & 24.07 & 43.46 & 71.86 & $>$100,000 \\
& Qwen3-30B~\cite{yang2025qwen3} & 23.37 & 55.96 & 81.16 & 30,000 & 14.52 & 36.56 & 78.07 & 30,000 & 24.81 & 42.84 & 70.80 & 30,000 \\
& Qwen3-30B~\cite{yang2025qwen3} + PKG & 23.31 & 56.15 & 81.06 & 30,000 & 14.63 & 36.53 & 78.11 & 30,000 & 25.19 & 43.95 & 71.98 & 30,000 \\
& PKG beam search & 22.38 \scriptsize $\pm$ 0.26 & 55.74 \scriptsize $\pm$ 0.25 & 80.92 \scriptsize $\pm$ 0.26 & 41.87 & 13.32 \scriptsize $\pm$ 0.34 & 37.42 \scriptsize $\pm$ 1.19 & 78.93 \scriptsize $\pm$ 2.06 & 42.90 & 24.96 \scriptsize $\pm$ 1.93 & 43.46 \scriptsize $\pm$ 2.42 & 72.18 \scriptsize $\pm$ 0.55 & 41.74 \\
& PDPP~\cite{wang2023pdpp} & 36.73 \scriptsize $\pm$ 0.59 & 61.96 \scriptsize $\pm$ 0.59 & 83.20 \scriptsize $\pm$ 0.33 & 41.87 & 22.37 \scriptsize $\pm$ 0.57 & 44.60 \scriptsize $\pm$ 0.16 & 83.00 \scriptsize $\pm$ 0.42 & 42.90 & 26.52 \scriptsize $\pm$ 1.56 & 45.58 \scriptsize $\pm$ 1.85 & \textbf{74.89} \scriptsize $\pm$ 0.85 & 41.74 \\
& KEPP~\cite{nagasinghe2024not} & 34.93 \scriptsize $\pm$ 2.60 & 60.34 \scriptsize $\pm$ 1.61 & 82.67 \scriptsize $\pm$ 0.69 & 42.18 & 13.85 \scriptsize $\pm$ 7.49 & 28.40 \scriptsize $\pm$ 12.26 & 62.54 \scriptsize $\pm$ 14.35 & 44.66 & 27.56 \scriptsize $\pm$ 1.48 & \underline{45.93} \scriptsize $\pm$ 2.37 & \underline{74.36} \scriptsize $\pm$ 0.97 & 41.86 \\
& PlanLLM~\cite{yang2025planllm} & 36.84 \scriptsize $\pm$ 1.21 & 61.56 \scriptsize $\pm$ 1.03 & 83.23 \scriptsize $\pm$ 0.53 & 384.94 & \underline{33.44} \scriptsize $\pm$ 0.15 & \textbf{51.05} \scriptsize $\pm$ 0.46 & \textbf{84.66} \scriptsize $\pm$ 0.41 & 386.43 & \underline{30.00} \scriptsize $\pm$ 1.41 & 44.35 \scriptsize $\pm$ 2.52 & 73.60 \scriptsize $\pm$ 1.66 & 384.77 \\
& SCHEMA~\cite{niu2024schema} & \underline{37.24} \scriptsize $\pm$ 0.60 & \underline{62.69} \scriptsize $\pm$ 0.28 & \textbf{83.94} \scriptsize $\pm$ 0.23 & 6.13 & 32.89 \scriptsize $\pm$ 0.61 & 50.84 \scriptsize $\pm$ 0.47 & \underline{83.98} \scriptsize $\pm$ 0.67 & 6.28 & 26.30 \scriptsize $\pm$ 1.49 & 42.77 \scriptsize $\pm$ 2.12 & 73.04 \scriptsize $\pm$ 1.42 & 6.12 \\
\highlightrow
& ViterbiPlanNet & \textbf{38.45} \scriptsize $\pm$ 0.32 & \textbf{63.07} \scriptsize $\pm$ 0.17 & \underline{83.89} \scriptsize $\pm$ 0.16 & 5.57 & \textbf{33.99} \scriptsize $\pm$ 0.23 & \underline{50.87} \scriptsize $\pm$ 0.17 & 83.88 \scriptsize $\pm$ 0.31 & 6.67 & \textbf{32.37} \scriptsize $\pm$ 0.96 & \textbf{46.96} \scriptsize $\pm$ 1.75 & 73.85 \scriptsize $\pm$ 0.85 & 5.48 \\
\rowcolor{gray!10}
& Improvement & \colorbox{yellow!30}{+1.21 \scriptsize $\pm$ 0.69} & \colorbox{yellow!30}{+0.38 \scriptsize $\pm$ 0.34} & -0.05 \scriptsize $\pm$ 0.27 &  & \colorbox{yellow!30}{+0.55 \scriptsize $\pm$ 0.27} & -0.18 \scriptsize $\pm$ 0.49 & \colorbox{yellow!30}{-0.78 \scriptsize $\pm$ 0.50} &  & \colorbox{yellow!30}{+2.37 \scriptsize $\pm$ 1.63} & +1.04 \scriptsize $\pm$ 3.06 & \colorbox{yellow!30}{-1.04 \scriptsize $\pm$ 1.22} &  \\
\midrule
\multirow{11}{*}{4} 
& Qwen2.5-VL-32B~\cite{bai2025qwen2} & 5.56 & 31.22 & 66.31 & 32,000 & 1.87 & 17.05 & 55.66 & 32,000 & 5.26 & 28.84 & 60.21 & 32,000 \\
& Qwen2.5-32B~\cite{qwen2024qwen2} & 9.22 & 46.32 & 76.15 & 32,000 & 4.98 & 27.45 & 71.64 & 32,000 & 23.25 & 41.89 & 73.91 & 32,000 \\
& Gemini 2.5 Pro~\cite{comanici2025gemini} & 14.00 & 51.33 & 78.58 & $>$100,000 & 8.10 & 31.90 & 71.70 & $>$100,000 & 22.37 & 40.35 & 73.05 & $>$100,000 \\
& Qwen3-30B~\cite{yang2025qwen3} & 10.59 & 49.06 & 78.03 & 30,000 & 4.64 & 28.85 & 70.45 & 30,000 & 22.37 & 41.23 & 73.90 & 30,000 \\
& Qwen3-30B~\cite{yang2025qwen3} + PKG & 10.96 & 48.77 & 77.48 & 30,000 & 4.78 & 29.00 & 71.04 & 30,000 & 21.93 & 41.67 & 74.43 & 30,000 \\
& PKG beam search & 9.30 \scriptsize $\pm$ 0.22 & 47.65 \scriptsize $\pm$ 0.54 & 78.25 \scriptsize $\pm$ 0.42 & 41.87 & 5.14 \scriptsize $\pm$ 0.60 & 31.29 \scriptsize $\pm$ 3.64 & 74.26 \scriptsize $\pm$ 5.38 & 42.90 & 21.23 \scriptsize $\pm$ 0.96 & 40.86 \scriptsize $\pm$ 0.83 & 72.69 \scriptsize $\pm$ 0.75 & 41.74 \\
& PDPP~\cite{wang2023pdpp} & 21.47 \scriptsize $\pm$ 2.09 & 55.66 \scriptsize $\pm$ 1.64 & 80.68 \scriptsize $\pm$ 0.83 & 41.87 & 15.21 \scriptsize $\pm$ 0.34 & 41.01 \scriptsize $\pm$ 0.32 & 81.64 \scriptsize $\pm$ 0.48 & 42.90 & 21.40 \scriptsize $\pm$ 0.53 & 40.20 \scriptsize $\pm$ 2.00 & 72.82 \scriptsize $\pm$ 1.84 & 41.74 \\
& KEPP~\cite{nagasinghe2024not} & 22.34 \scriptsize $\pm$ 0.43 & 55.24 \scriptsize $\pm$ 0.30 & 80.58 \scriptsize $\pm$ 0.25 & 42.18 & 15.20 \scriptsize $\pm$ 1.27 & 33.39 \scriptsize $\pm$ 0.73 & 67.79 \scriptsize $\pm$ 1.29 & 44.66 & 22.54 \scriptsize $\pm$ 1.93 & \underline{42.46} \scriptsize $\pm$ 1.49 & 73.11 \scriptsize $\pm$ 0.94 & 41.86 \\
& PlanLLM~\cite{yang2025planllm} & 22.91 \scriptsize $\pm$ 1.39 & 55.29 \scriptsize $\pm$ 1.54 & 81.03 \scriptsize $\pm$ 0.47 & 384.94 & \underline{23.19} \scriptsize $\pm$ 0.32 & \textbf{45.70} \scriptsize $\pm$ 0.33 & \textbf{83.44} \scriptsize $\pm$ 0.39 & 386.43 & 23.42 \scriptsize $\pm$ 1.40 & 41.95 \scriptsize $\pm$ 2.81 & 72.32 \scriptsize $\pm$ 0.91 & 384.77 \\
& SCHEMA~\cite{niu2024schema} & \underline{24.18} \scriptsize $\pm$ 0.47 & \textbf{57.02} \scriptsize $\pm$ 0.64 & \textbf{81.46} \scriptsize $\pm$ 0.19 & 6.13 & 22.33 \scriptsize $\pm$ 0.92 & 45.21 \scriptsize $\pm$ 1.05 & \underline{82.93} \scriptsize $\pm$ 0.25 & 6.28 & \underline{24.39} \scriptsize $\pm$ 1.84 & 41.14 \scriptsize $\pm$ 3.62 & \underline{73.13} \scriptsize $\pm$ 1.97 & 6.12 \\
\highlightrow
& ViterbiPlanNet & \textbf{24.64} \scriptsize $\pm$ 0.30 & \underline{57.00} \scriptsize $\pm$ 0.42 & \underline{81.18} \scriptsize $\pm$ 0.44 & 5.60 & \textbf{23.92} \scriptsize $\pm$ 0.29 & \underline{45.63} \scriptsize $\pm$ 0.55 & 82.56 \scriptsize $\pm$ 0.44 & 6.87 & \textbf{27.54} \scriptsize $\pm$ 0.70 & \textbf{45.55} \scriptsize $\pm$ 1.89 & \textbf{74.71} \scriptsize $\pm$ 1.19 & 5.50 \\
\rowcolor{gray!10}
& Improvement & +0.46 \scriptsize $\pm$ 0.61 & -0.02 \scriptsize $\pm$ 0.78 & -0.29 \scriptsize $\pm$ 0.49 &  & \colorbox{yellow!30}{+0.73 \scriptsize $\pm$ 0.44} & -0.08 \scriptsize $\pm$ 0.62 & \colorbox{yellow!30}{-0.88 \scriptsize $\pm$ 0.59} &  & \colorbox{yellow!30}{+3.15 \scriptsize $\pm$ 1.93} & \colorbox{yellow!30}{+3.09 {\scriptsize $\pm$ 2.43}} & +1.58 \scriptsize $\pm$ 2.37 &  \\
\bottomrule
\end{tabular}%
}
\end{table*}

\begin{table}[t]
\centering
\caption{Comparison with MTID$^{\clubsuit}$ on CrossTask.}
\label{tab:performance_comparison_MTID_crosstask}
\resizebox{\linewidth}{!}{%
\begin{tabular}{@{}ll cccr@{}}
\toprule
\multirow{2}{*}{\textbf{Horizon}} & \multirow{2}{*}{\textbf{Method}} & \multicolumn{4}{c}{\textbf{CrossTask}} \\
\cmidrule(lr){3-6}
& & SR $\uparrow$ (\%) & mAcc $\uparrow$ (\%) & mIoU $\uparrow$ (\%) & Params (M) \\
\midrule
\multirow{2}{*}{T = 3} 
& MTID$^{\clubsuit}$~\cite{zhou2025masked} & \textbf{40.45} & \underline{67.19} & \underline{69.17} & 1,085.20 \\
& ViterbiPlanNet$^{\clubsuit}$ & \underline{39.75} & \textbf{67.39} & \textbf{76.92} & 5.49 \\
\midrule
\multirow{2}{*}{T = 4} 
& MTID$^{\clubsuit}$~\cite{zhou2025masked} & \textbf{24.76} & \underline{60.69} & \underline{67.67} & 1,085.20 \\
& ViterbiPlanNet$^{\clubsuit}$ & \underline{24.19} & \textbf{61.12} & \textbf{80.67} & 5.52 \\
\bottomrule
\end{tabular}%
}
\end{table}

\subsection{Comparisons with the State of the Art}
\paragraph{Compared Approaches.} We compare ViterbiPlanNet against recent state-of-the-art methods for procedure planning in instructional videos, spanning diffusion-based planning (KEPP~\cite{nagasinghe2024not}, PDPP~\cite{wang2023pdpp}, MTID~\cite{zhou2025masked}), LLM-derived structured memory (SCHEMA~\cite{niu2024schema}), and multimodal commonsense reasoning (PlanLLM~\cite{yang2025planllm}).
We also assess the performance of a classic method that uses procedural knowledge with the
baseline termed \textit{PKG beam search}. Here, we first train a step model $f_{step}$ based on $f_{enc}$ to predict start and end actions, and a beam search is then applied directly over the PKG.
Finally, we introduce a series of baselines where large language (LLMs) and vision-language (VLMs) models tackle planning via in-context learning.
LLMs are provided with start/end actions predicted by $f_{step}$, the action taxonomy, and demonstration trajectories sampled from the training set, while VLMs receive start/goal frames with the same context. We evaluate Gemini 2.5 Pro~\cite{comanici2025gemini}, Qwen3-30B~\cite{yang2025qwen3}, Qwen2.5-32B~\cite{qwen2024qwen2}, and Qwen2.5-VL-32B~\cite{bai2025qwen2}, along with a PKG-augmented variant of Qwen3-30B where the graph is injected into the prompt following~\cite{fatemi2024talk}. These comparisons are meant to assess %
visual planning abilities of current LLMs when compared with trained methods\footnote{See supplementary material for more details.\label{fn:supp9}}.

\paragraph{Performance on Different Planning Horizons.}
Table~\ref{tab:performance_combined_viterbiplannet} reports a comparison across CrossTask, COIN, and NIV for horizons $T \in \{3,4\}$. Results for $T \in \{5,6\}$ are in the supplementary material. We observe these trends:\\
\ul{ViterbiPlanNet achieves the highest Success Rate (SR)} across all settings, with statistically significant improvements over prior methods ({\setlength{\fboxsep}{0pt}\colorbox{yellow!30}{marked in yellow}}). Performance is comparable to second-best alternatives (either SCHEMA or PlanLLM) according to mAcc and mIoU on CrossTask and COIN, with either small decrements or statistically inconclusive performance differences, suggesting that despite prioritizing global consistency, ViterbiPlanNet maintains robust step-level accuracy. Improvements are positive and statistically significant on NIV according to all metrics. ViterbiPlanNet's leading performance in Success Rate (the most stringent metric) on all datasets is due to the ability of our method to perform sequential modeling, hence attaining more correct sequences than competitors, influencing SR. 
\\
\ul{In-context LLM/VLM models achieve limited performance.} Vision-only prompting (Qwen2.5-VL-32B) performs particularly poorly, confirming that current VLMs struggle to infer complex multi-step procedures directly from visual inputs. Computing start/end actions with a step model and allowing LLMs to reason symbolically improves results (Qwen2.5-32B and Qwen3-30B). Adding the PKG in the prompt does not improve results (Qwen2.5-VL-32B + PKG), highlighting that current models struggle to incorporate structured procedural knowledge.
Even the strongest foundation model Gemini~2.5~Pro attains results far behind training-based methods, showing that prompting alone is insufficient for structured instructional reasoning with LLMs. 
Notably, the simple PKG beam-search baseline outperforms most LLM/VLMs, demonstrating the value of explicit procedural priors.%

\paragraph{Comparison with MTID.} Table~\ref{tab:performance_comparison_MTID_crosstask} compares ViterbiPlanNet to MTID~\cite{zhou2025masked} on CrossTask$^{\ref{fn:supp9}}$.
Since re-training MTID is computationally prohibitive (more than a billion parameters), we report available results from the original paper. Note that MTID adopts different evaluation settings, training data, and features, so we conform ViterbiPlanNet to this protocol for fair comparison, and mark results with $^{\clubsuit}$.
Despite the significant parameter count difference (1B vs 5M), ViterbiPlanNet remains competitive even in these settings, achieving superior mIoU, and comparable SR and mAcc.

\begin{figure}[t]
    \centering
    \includegraphics[width=0.49\linewidth]{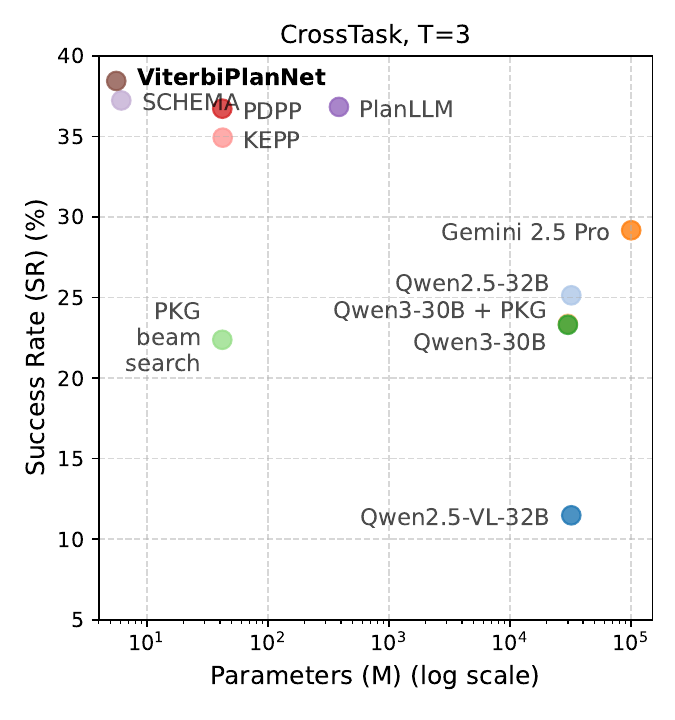}
    \hfill
    \includegraphics[width=0.49\linewidth]{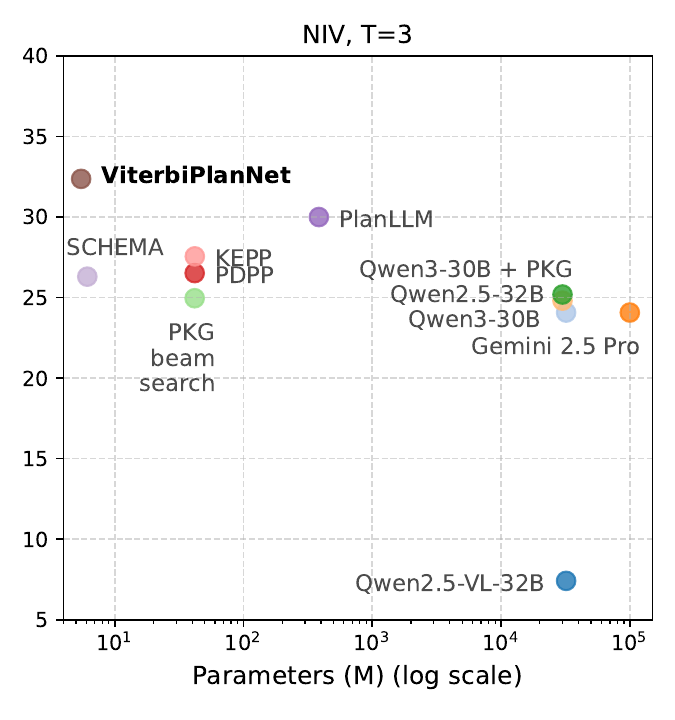}
    \caption{Parameter Efficiency on CrossTask and NIV.}
    \label{fig:params}
\end{figure}

\paragraph{Parameter Efficiency.}
Fig.~\ref{fig:params} shows the number of parameters (M, log-scale) versus SR on CrossTask and NIV for $T{=}3$ across all methods. 
As illustrated in the figure and also detailed in Table~\ref{tab:performance_combined_viterbiplannet}, ViterbiPlanNet is the most parameter-efficient, operating with only $\sim$5--7M parameters—two to three orders of magnitude fewer than competing approaches such as language models ($\sim 30B-100B$), MTID ($1.08B$) and PlanLLM ($\sim385M$). This efficiency stems from the simplicity of $f_{emiss}$ and the fact that ViterbiPlanNet does not need to memorize procedural knowledge, saving parameters. Among competitors, SCHEMA has similar parameter counts but lower performance.

\begin{table}[t]
\vspace{-9pt}
\centering
\caption{Cross-Horizon Consistency results on CrossTask.}
\label{tab:horizon_comparison}
\resizebox{\linewidth}{!}{%
\begin{tabular}{l l l l}
\toprule
\textbf{Method} & \textbf{SR} $\uparrow$ \textbf{(\%) [6 $\rightarrow$ 3]} & \textbf{SR} $\uparrow$ \textbf{(\%) [6 $\rightarrow$ 4]} & \textbf{SR} $\uparrow$ \textbf{(\%) [6 $\rightarrow$ 5]} \\
\midrule
Qwen2.5-VL-32B~\cite{bai2025qwen2} & 10.38 & 6.35 & 2.59 \\
Qwen2.5-32B~\cite{qwen2024qwen2} & 16.88 & 7.24 & 4.28 \\
Gemini 2.5 Pro~\cite{comanici2025gemini} & \underline{20.97} & \underline{10.46} & 4.57 \\
Qwen3-30B~\cite{yang2025qwen3} & 17.17 & 8.14 & 4.61 \\
Qwen3-30B~\cite{yang2025qwen3} + PKG & 17.83 & 8.29 & 4.06 \\
PKG beam search & 18.31 \scriptsize $\pm$ 0.40 & 8.26 \scriptsize $\pm$ 0.18 & 5.14 \scriptsize $\pm$ 0.22 \\
\midrule
PDPP~\cite{wang2023pdpp} & 12.95 \scriptsize $\pm$ 0.55 & 6.58 \scriptsize $\pm$ 0.36 & 2.80 \scriptsize $\pm$ 0.53 \\
KEPP~\cite{nagasinghe2024not} & 12.05 \scriptsize $\pm$ 0.54 & 7.11 \scriptsize $\pm$ 1.24 & \underline{6.39} \scriptsize $\pm$ 1.13 \\
PlanLLM~\cite{yang2025planllm} & 10.55 \scriptsize $\pm$ 0.60 & 5.96 \scriptsize $\pm$ 0.79 & 2.73 \scriptsize $\pm$ 0.19 \\
SCHEMA~\cite{niu2024schema} & 16.12 \scriptsize $\pm$ 1.24 & 9.69 \scriptsize $\pm$ 0.90 & 5.67 \scriptsize $\pm$ 0.73 \\
\textbf{ViterbiPlanNet} & \textbf{27.77} \scriptsize $\pm$ 0.43 & \textbf{18.45} \scriptsize $\pm$ 0.39 & \textbf{10.21} \scriptsize $\pm$ 0.50 \\
\midrule
Improvement & +6.80 \scriptsize $\pm 0.43$ & +7.99 \scriptsize $\pm 0.39$ & +3.82 \scriptsize $\pm 1.28$ \\
\bottomrule
\end{tabular}%
}
\end{table}

\paragraph{Cross-Horizon Consistency.}
The standard protocol for procedure understanding prescribes to train different models for specific planning horizons $T$. We argue that a robust procedural planner should predict plans which are coherent across different horizons. To assess such robustness, we propose a new protocol where models are trained with a long horizon $T=6$ and evaluated on shorter horizons $T \in \{3,4,5\}$. Since every length-6 trajectory contains shorter subsequences, this protocol tests whether models truly learn procedural planning %
rather than memorizing horizon-specific patterns. We report results on CrossTask in Table~\ref{tab:horizon_comparison}. Among competitors, LLMs and VLMs (e.g., Gemini~2.5~Pro, Qwen variants) show moderate robustness, whereas learning-based approaches such as PDPP, KEPP, PlanLLM, and SCHEMA generally struggle to output consistent plans. In contrast, the proposed ViterbiPlanNet exhibits a much greater robustness to mismatches between training and testing horizons, %
with significant margins across metrics. These results highlight the effectiveness of the Differentiable Viterbi Layer (DVL) in capturing transferable procedural structure, enabling cross-horizon consistency, with improvements of up to $\approx 8\%$ with respect to competitors.

\begin{figure}
    \centering
    \includegraphics[width=0.95\linewidth]{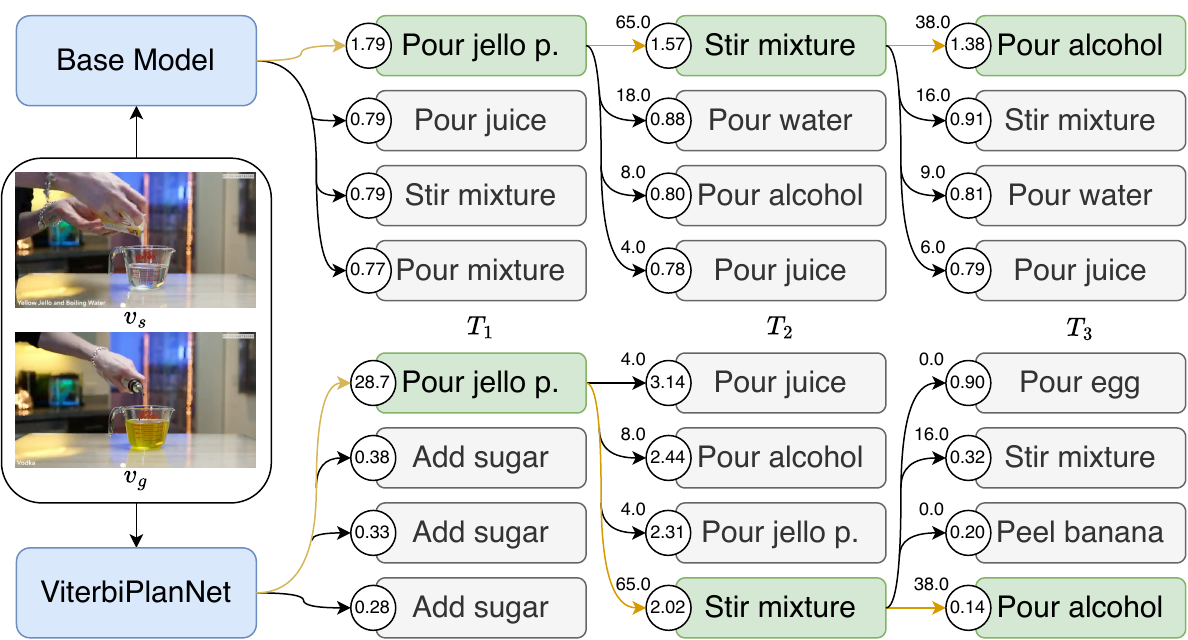}
    \caption{Qualitative comparison. The Base Model (top) learns to implicitly memorize the PKG, baking transition probabilities (arrows) directly into its predictions (circles). ViterbiPlanNet (bottom) learns smoother emissions decoupled from the graph, relying on the PKG's structural guidance for decoding.}
    \label{fig:qualitative_result}
    \vspace{-10pt}
\end{figure}

\paragraph{Qualitative analysis.} 
Fig.~\ref{fig:qualitative_result} compares the probability distributions of the Base Model (conf. 1 in Table~\ref{tab:ablation_reorg}) and the emission probabilities produced by ViterbiPlanNet. 
Predictions in the Base Model are aligned to transition probabilities, suggesting implicit memorization of the graph. On the contrary, emissions predicted by ViterbiPlanNet are decoupled from transition probabilities. This makes it easier to learn emissions (no need to rule out ``pour egg'' at $T=3$ as \texttt{stir mixture} $\to$ \texttt{pour egg} is impossible), and also leave more room for alternative decodings. For instance, \texttt{Pour Jello Powder}, \texttt{Stir Mixture}, \texttt{Stir Mixture} is both reasonable and possible under our model.

\vspace{-5pt}
\section{Conclusion}
This work demonstrates that explicitly integrating procedural knowledge end-to-end is an effective strategy for video procedure planning. Our method, ViterbiPlanNet, uses a Differentiable Viterbi Layer with a Procedure Knowledge Graph to learn emission probabilities instead of full plans. We show that this structure-aware training is substantially more effective than using Viterbi solely for post-processing. Consequently, ViterbiPlanNet is highly parameter- and sample-efficient, outperforming competitors on three datasets and showing robust cross-horizon consistency. We also introduced a new unified evaluation protocol to assess progress more robustly.
We believe our work will encourage interest in the inclusion of structural priors at training time, advancing efficient on-device planning for future assistive agents.

\section{Acknowledgments}
This research is supported in part by the PNRR PhD scholarship ``Digital Innovation: Models, Systems and Applications'' DM 118/2023, by the project Future Artificial Intelligence Research (FAIR) – PNRR MUR Cod. PE0000013 - CUP: E63C22001940006, and by the Research Program PIAno di inCEntivi per la Ricerca di Ateneo 2020/2022 — Linea di Intervento 3 ``Starting Grant'' EVIPORES Project - University of Catania. We thank CINECA under the ISCRA initiative, for HPC resources and support. We also thank the University of Bath for the Visiting Postgraduate Scholars Scheme that allowed author Luigi Seminara to carry out this work while visiting the University of Bath.

{
   \small
   \bibliographystyle{ieeenat_fullname}
   \bibliography{main}
}

\cftsetindents{section}{0em}{2em}
\cftsetindents{subsection}{0.5em}{3em}

\doparttoc %
\faketableofcontents %

\clearpage
\maketitlesupplementary

\patchcmd{\numberline}{\hfil}{\hfil}{}{}

\setcounter{secnumdepth}{-2} %
\part{} %
\setcounter{secnumdepth}{2} %
\parttoc %

\addcontentsline{toc}{section}{Appendices}
\renewcommand{\thesubsection}{\thesection.\Alph{subsection}}

\newfloat{prompt}{htbp}{lop}
\floatname{prompt}{Prompt}

\begin{figure}[t]
    \centering
    \includegraphics[width=\linewidth]{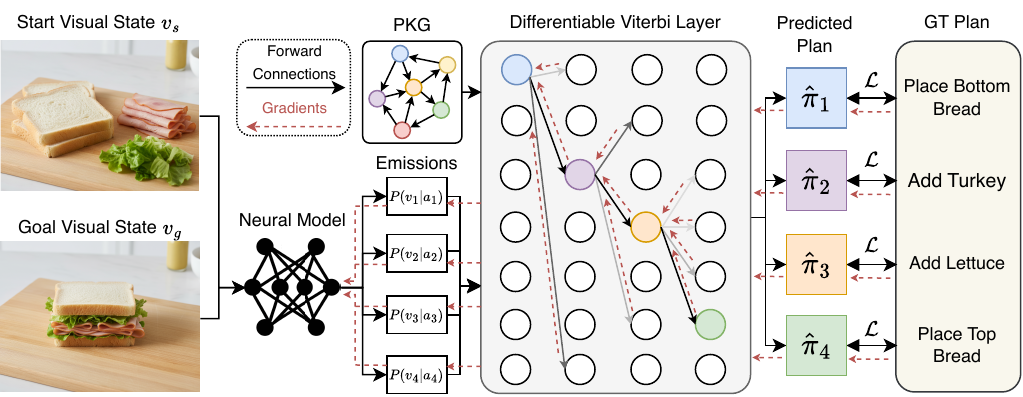}
    \caption{Given start and goal visual states, a {neural model} computes step-wise {emissions}. We propose a {Differentiable Viterbi Layer} that uses a {Procedural Knowledge Graph (PKG)} to decode emissions into a {predicted plan}. The layer allows {gradients} from the planning {Loss ($\mathcal{L}$)} to flow and train the neural model end-to-end, forcing it to learn structure-aware visual representations.}
    \label{fig:teaser_supp}
\end{figure}

In this supplementary material we provide more details about our method (see Figure~\ref{fig:teaser_supp}) and the experimental setup. We point to each footnote in the main paper with the following notation: ($fn\ x$) where $x$ is the footnote number in the main paper.

\section{Method}
\label{appendix:method}
\subsection{Problem Formulation ($fn\ 1$)}
\label{appendix:problem_formulation}

\paragraph{Marginalizing the initial latent action \(a_0\).}
We consider a latent action sequence \(a_{0:T} = (a_0, a_1, \dots, a_T)\) and a corresponding sequence of observed visual states \(v_{0:T} = (v_s = v_0, v_1, \dots, v_T = v_g)\). Under the first--order Markov assumption, the model is factorized into independent transition and emission components. The joint distribution over latent actions and visual observations is therefore:

\begin{equation}
\small
P(a_{0:T}, v_{0:T})
= P(a_0)\,P(v_0 \mid a_0)
  \prod_{t=1}^{T} P(a_t \mid a_{t-1})\,P(v_t \mid a_t).
\label{eq:joint}
\end{equation}

Our objective is to infer the posterior distribution over the latent plan
\(\pi = a_{1:T}\) given the full sequence of visual observations. Using Bayes' rule and marginalizing the unobserved initial action \(a_0\), we obtain:
\begin{align}
\small
&P(a_{1:T} \mid v_{0:T})
= \frac{\sum_{a_0 \in \mathcal{K}} P(a_{0:T}, v_{0:T})}{P(v_{0:T})} \nonumber\\[2pt]
&\propto \sum_{a_0 \in \mathcal{K}}
P(a_0)\,P(v_0 \mid a_0)
\prod_{t=1}^{T} P(a_t \mid a_{t-1})\,P(v_t \mid a_t),
\label{eq:posterior_marg}
\end{align}
where \(\mathcal{K}\) denotes the discrete action set. Only the terms involving \(a_0\) depend on the marginalization, and all other factors remain unaffected. It is therefore convenient to group the contributions of \(a_0\) into an \emph{effective prior} over the first action \(a_1\). For any action class \(K_j \in \mathcal{K}\), we thus have:

\begin{equation}
\small
P(a_1 = K_j \mid v_0)
\;\propto\;
\sum_{a_0 \in \mathcal{K}}
P(a_0)\,P(v_0 \mid a_0)\,P(a_1 = K_j \mid a_0).
\label{eq:effective_prior}
\end{equation}

However, in practice, the quantity \(P(a_0)\,P(v_0 \mid a_0)\) cannot be estimated directly because the initial action is unobserved and the model lacks supervision for this term. Following common practice, we thus assume a \emph{uniform prior} over the first action \(a_1\). Under this assumption, all structural information at \(t=1\) is captured by the emission term \(P(v_1 \mid a_1)\).
Substituting this simplification into Eq.~\eqref{eq:posterior_marg}, the posterior becomes:

\begin{align}
\small
P(a_{1:T} \mid v_{0:T})
\;\propto\;
\prod_{t=1}^{T} P(a_t \mid a_{t-1})\,P(v_t \mid a_t),
\label{eq:posterior_final}
\end{align}
which is the quantity whose maximization yields the most probable latent plan.

\subsection{Viterbi Algorithm}
\label{appendix:viterbi_algorithm}

The Viterbi algorithm~\cite{1054010} is a dynamic programming method for computing the 
most likely sequence of latent states in a Hidden Markov Model (HMM).
Given (1) a set of $N$ discrete states $\mathcal{K}=\{K_1,\dots,K_N\}$,
(2) transition probabilities $P(a_t \mid a_{t-1})$, and 
(3) emission probabilities $P(v_t \mid a_t)$,
the goal is to find the most probable sequence of hidden actions
$a_{1:T}$ that explains the observations $v_{1:T}$.

\paragraph{Objective.}
The algorithm maximizes the posterior Eq.~\eqref{eq:posterior_final} as follows:
\begin{equation}
    \pi^*
    = \underset{\pi =a_{1:T} \in \mathcal{K}^T}{\arg\max} \; \prod_{t=1}^T \underbrace{P(a_t | a_{t-1})}_{\text{Transition}} \underbrace{P(v_{t}| a_t)}_{\text{Emission}}.
    \label{eq:viterbi_objective}
\end{equation}

\paragraph{Dynamic programming recursion.}
To efficiently compute~\eqref{eq:viterbi_objective}, Viterbi stores: (1) \emph{state scores} $\delta_t(j)$, the best score of any path ending in state $K_j$ at time $t$; (2) \emph{backpointers} $\psi_t(j)$, the most likely predecessor of $K_j$.
Since no predecessor exists at $t=1$, initial scores depend only on emissions:
\begin{equation}
\small
\delta_1(j) = P(v_1 \mid a_1 = K_j), 
\qquad j = 1,\dots,N.
\label{eq:viterbi_init}
\end{equation}
 For each time step $t>1$ and each state $K_j$, Viterbi computes:
\begin{align}
\small
\delta_t(j) = & \nonumber\\
&P(v_t \mid a_t=K_j) \cdot \nonumber\\
&\max_{i \in \{1,\dots,N\}}
\left[
\delta_{t-1}(i)\,P(a_t=K_j \mid a_{t-1}=K_i)
\right],
\label{eq:viterbi_delta} \\
\psi_t(j) = &
\underset{{i \in \{1,\dots,N\}}}{\arg\max}
\left[
\delta_{t-1}(i)\,P(a_t=K_j \mid a_{t-1}=K_i)
\right].
\label{eq:viterbi_psi}
\end{align}
The $\max$ operator ensures that only the best predecessor state contributes to the score.

\paragraph{Backtracking.}
Once $\delta_T$ is computed, the final state is chosen as follows:
\begin{equation}
\small
a_T^* = \arg\max_j \delta_T(j),
\end{equation}
and the full optimal plan is reconstructed by tracing back the stored pointers:
\begin{equation}
\small
a_t^* = \psi_{t+1}(a_{t+1}^*), 
\qquad t=T-1,\dots,1.
\end{equation}

\paragraph{Interpretation.}
The Viterbi algorithm guarantees:
\begin{itemize}
    \item \textbf{Optimality}: the returned sequence maximizes the posterior probability;
    \item \textbf{Efficiency}: complexity $O(TN^2)$ instead of exponential search $O(N^T)$;
    \item \textbf{Modularity}: transitions and emissions contribute separately, matching the probabilistic factorization used in Eq.~\eqref{eq:viterbi_objective}.
\end{itemize}

\paragraph{Connection to our formulation.}
In our framework emissions are predicted from visual encodings, and transitions come from the Procedural Knowledge Graph (PKG). This correspondence makes Viterbi a natural decoding algorithm for procedural planning. However, its \(\max\) and \(\arg\max\) operations prevent end-to-end learning, motivating the introduction of our Differentiable Viterbi Layer (DVL), which replaces these operators with smooth relaxations following~\cite{mensch2018differentiable}.

\subsection{Differentiable Viterbi ($fn\ 5$)}
\label{appendix:differentiable_viterbi}
\paragraph{Smooth max and soft argmax operators.}
Let $\mathbf{x} \in \mathbb{R}^N$ be a generic score vector, we define the following \emph{smooth max} (log-sum-exp) and \emph{soft argmax} (softmax) operations which aim to extract from $\mathbf{x}$ a max-like value and a distribution over indexes corresponding to entries closer to the maximum value in a differentiable fashion:
\begin{align}
\small
   &\mathrm{S\text{-}max}(\mathbf{x}) = \log\left(\sum_{k=1}^{N} \exp(x_k - m)\right) + m \\
   &\mathrm{S\text{-}argmax}(\mathbf{x})_k = \frac{\exp(x_k - m)}{\sum_{j=1}^{N} \exp(x_j - m)}
 \end{align}
where $m=\max (\mathbf{x})$. Intuitively, $\mathrm{S\text{-}max}(\mathbf{x})$ returns a value close to $m$, while remaining differentiable in all components of  $\mathbf{x}$. The $\mathrm{S\text{-}argmax}$ corresponds to the standard softmax function, producing a probability distribution over indices that reflects their relative proximity to the maximum.

\paragraph{Differences with respect to \cite{mensch2018differentiable}.}
Our Differentiable Viterbi Layer (DVL) builds on the general framework of differentiable dynamic programming proposed by Mensch and Blondel~\cite{mensch2018differentiable}, but differs in several important respects. First, while \cite{mensch2018differentiable} introduced a unified approach to smoothing dynamic programs and enabled end-to-end training of both transition and emission potentials, our DVL does not introduce new trainable parameters: transition probabilities are fixed by the procedural knowledge graph (PKG), and emission probabilities are provided by upstream modules. Second, we explicitly introduce \emph{soft backpointer distributions} and their recursive composition into a \emph{soft plan}, which serves as a differentiable analogue of the discrete Viterbi backtrace. In summary, our contribution is a re-designed decoding-only layer that leverages fixed structural knowledge to produce differentiable plans, enabling gradient flow through decoding without learning dynamic programming parameters.

\subsection{Visual Encoding ($fn\ 5$)}
\label{appendix:visual_encoding}
Following prior work, we employ S3D~\cite{xie2018rethinking} as our visual backbone to extract spatiotemporal features from start and goal video states. The resulting representations are then passed through a learned projection layer, which adapts the backbone features to the dimensionality required by our planning model.

\subsection{Training ($fn\ 6$)}
\label{appendix:training}
ViterbiPlanNet is trained end-to-end by minimizing a composite loss function $\mathcal{L}$ defined as a sum of three distinct loss components with equal weights.

\paragraph{Visual-Semantic Alignment Loss ($\mathcal{L}_{\text{align}}$).}
To encourage the model to learn a structured state space, we align visual representations with textual descriptions of their corresponding procedural states as in~\cite{niu2024schema}. Following the idea of SCHEMA~\cite{niu2024schema}, for each action in the vocabulary we obtain a set of natural language descriptions of its \emph{before-state} and \emph{after-state} using the same prompt and model used in SCHEMA~\cite{niu2024schema}.
We denote this set of descriptions as $\mathcal{A}$. These descriptions capture discriminative object attributes (e.g., ``the pan has no onion on it'' before \textit{add onion}). 

Formally, given the encoded start state $v_s^{enc}$, the goal state $v_g^{enc}$, and a textual description $d \in \mathcal{A}$ encoded by a frozen language model into an embedding $d^{enc}$, we compute cosine similarities between the visual embeddings and all candidate textual descriptions:
\begin{align}
\small
    \text{sim}(v_s^{enc}, d^{enc}) = \frac{v_s^{enc} \cdot d^{enc}}{\|v_s^{enc}\|\|d^{enc}\|}, \\
    \text{sim}(v_g^{enc}, d^{enc}) = \frac{v_g^{enc} \cdot d^{enc}}{\|v_g^{enc}\|\|d^{enc}\|}.
\end{align}
During training, the \emph{positive samples} are the textual descriptions corresponding to the ground-truth first action’s before-state (for $v_s$) and the ground-truth last action’s after-state (for $v_g$). All other descriptions act as negatives. We adopt a contrastive cross-entropy loss as in~\cite{niu2024schema}, which encourages the visual embeddings to be close to their correct textual descriptions while being far from incorrect ones:
\begin{align}
\small
    \mathcal{L}_{align} &= \nonumber\\ 
    &- \log \frac{\exp(sim(v_s^{enc}, d_+^{enc}))}{\sum_{d \in \mathcal{A}} \exp(sim(v_s^{enc}, d^{enc}))} \nonumber\\
    &- \log \frac{\exp(sim(v_g^{enc}, d_+^{enc}))}{\sum_{d \in \mathcal{A}} \exp(sim(v_g^{enc}, d^{enc}))}.
\end{align}
where $d_+^{enc}$ denotes the encoding of the positive samples.  
This objective explicitly grounds the visual encoder in the semantics of procedural states, ensuring that the learned visual features capture the causal state changes relevant to the procedure.

\paragraph{Task Classification Loss ($\mathcal{L}_{\text{task}}$).}
Let $N_{tasks}$ denote the total number of possible tasks in the dataset.
We represent the ground-truth task label as a one-hot vector 
$c \in \{0,1\}^{N_{tasks}}$, 
where $c_{n} = 1$ if the procedure belongs to class $n$ and $0$ otherwise. 
As shown in the architecture, the auxiliary \textit{Task Head} takes the encoded visual features 
$(v_s^{\text{enc}}, v_g^{\text{enc}})$ as input and outputs a prediction vector 
$\hat c \in \mathbb{R}^{N_{tasks}}$, where $\hat c_{n}$ 
is the predicted score for class $n$ (for simplicity $\hat{c}$ is not labeled in the figure and is simply depicted as a square before $\mathcal{L}_{\textit{task}}$). This auxiliary prediction provides contextual information that implicitly guides the planning process. 

To train the Task Head, we minimize the Mean Squared Error (MSE) between the predicted 
scores and the one-hot ground-truth labels:
\begin{equation}
\small
    \mathcal{L}_{task} = \frac{1}{N_{tasks}} 
    \sum_{n=1}^{N_{tasks}} \big(\hat c_{n} - c_{n}\big)^2.
\end{equation}

\paragraph{Planning Loss ($\mathcal{L}_{\text{plan}}$).}
The central objective of our framework is to learn to generate the correct sequence of actions. The final output of the Structured Decoding module, $\tilde \pi \in \mathbb{R}^{T \times N}$, represents the refined score distribution over all $N$ possible actions at each of the $T$ time steps. We supervise these predictions against the one-hot encoded ground-truth plan $\tilde \pi^{GT} \in \{0,1\}^{T \times N}$, where $\tilde \pi^{GT}[t,n] = 1$ if the ground-truth action at time step $t$ is $K_n \in \mathcal{K}$ and $0$ otherwise. The Mean Squared Error (MSE) loss is then defined as:
\begin{equation}
    \mathcal{L}_{\text{plan}} 
    = \frac{1}{T} 
    \sum_{t=1}^{T} 
    \left( \tilde \pi_t - \tilde \pi^{GT}_t \right)^{2}.
    \label{eq:planning_loss_mse}
\end{equation}
Minimizing $\mathcal{L}_{\text{plan}}$ encourages the model to produce action distributions that closely match the target one-hot plan. With this the model learns accurate procedural sequences while being constrained by the structural priors encoded in the Procedural Knowledge Graph, which are enforced through the Differentiable Viterbi layers.

\paragraph{Overall Objective.}
The final training loss is a sum of the three components with equal weights:
\begin{equation}
    \mathcal{L} = \mathcal{L}_{plan} + \mathcal{L}_{align} + \mathcal{L}_{task}.
    \label{eq:total_loss_supp}
\end{equation}

\begin{table}[t]
\centering
\caption{Ablation of Viterbi components on CrossTask for $T \in \{4, 5, 6\}$.}
\label{tab:ablation_reorg_horizons}
\resizebox{0.97\linewidth}{!}{%

\begin{tabular}{
    c
    >{\centering\arraybackslash}p{0.9cm}  %
    >{\centering\arraybackslash}p{0.9cm}  %
    >{\centering\arraybackslash}p{0.9cm}  %
    r r r
}

\toprule
 & \multicolumn{1}{c}{\textbf{Train}} & \multicolumn{2}{c}{\textbf{Inference}} & \multicolumn{3}{c}{\textbf{Metrics (\%) $\uparrow$}} \\
\cmidrule(lr){2-2} \cmidrule(lr){3-4} \cmidrule(l){5-7}
 & \textbf{DVL} & \textbf{DVL} & \textbf{VD} & \multicolumn{1}{c}{\textbf{SR}} & \multicolumn{1}{c}{\textbf{mAcc}} & \multicolumn{1}{c}{\textbf{mIoU}} \\
\midrule
\multicolumn{7}{c}{\textbf{Horizon} $T=4$} \\
\midrule
1 & $\times$ & $\times$ & $\times$ & 18.93 \scriptsize $\pm$ 0.58 & 55.12 \scriptsize $\pm$ 0.46 & 79.87 \scriptsize $\pm$ 0.26 \\
2 & $\times$ & $\times$ & \checkmark & 18.64 \scriptsize $\pm$ 0.75 & 55.00 \scriptsize $\pm$ 0.38 & 79.78 \scriptsize $\pm$ 0.18 \\
3 & $\times$ & \checkmark & $\times$ & 21.54 \scriptsize $\pm$ 0.50 & 53.19 \scriptsize $\pm$ 0.46 & 79.90 \scriptsize $\pm$ 0.28 \\
4 & $\times$ & \checkmark & \checkmark & 19.92 \scriptsize $\pm$ 0.18 & 52.09 \scriptsize $\pm$ 0.43 & 79.78 \scriptsize $\pm$ 0.34 \\
\cmidrule(lr){1-7}
5 & \checkmark & $\times$ & $\times$ & 6.13 \scriptsize $\pm$ 0.31 & 44.71 \scriptsize $\pm$ 0.13 & 69.70 \scriptsize $\pm$ 0.56 \\
6 & \checkmark & $\times$ & \checkmark & 23.46 \scriptsize $\pm$ 0.20 & \textbf{57.13} \scriptsize $\pm$ 0.35 & \underline{81.05} \scriptsize $\pm$ 0.38 \\
7 & \checkmark & \checkmark & $\times$ & \underline{24.30} \scriptsize $\pm$ 0.66 & 56.42 \scriptsize $\pm$ 0.10 & 80.93 \scriptsize $\pm$ 0.48 \\
8 & \checkmark & \checkmark & \checkmark & \textbf{24.64} \scriptsize $\pm$ 0.30 & \underline{57.00} \scriptsize $\pm$ 0.42 & \textbf{81.18} \scriptsize $\pm$ 0.44 \\

\midrule
\multicolumn{4}{l}{Improvement w.r.t. conf. $1$} & 5.71 \scriptsize $\pm$ 0.64 & 1.88 \scriptsize $\pm$ 0.61 & 1.31 \scriptsize $\pm$ 0.52 \\

\midrule
\multicolumn{7}{c}{\textbf{Horizon} $T=5$} \\
\midrule
1 & $\times$ & $\times$ & $\times$ & 10.21 \scriptsize $\pm$ 0.08 & 50.49 \scriptsize $\pm$ 0.64 & 77.49 \scriptsize $\pm$ 0.44 \\
2 & $\times$ & $\times$ & \checkmark & 9.89 \scriptsize $\pm$ 0.08 & 50.44 \scriptsize $\pm$ 0.59 & 77.42 \scriptsize $\pm$ 0.44 \\
3 & $\times$ & \checkmark & $\times$ & 13.32 \scriptsize $\pm$ 0.26 & 48.64 \scriptsize $\pm$ 0.51 & 77.70 \scriptsize $\pm$ 0.37 \\
4 & $\times$ & \checkmark & \checkmark & 12.27 \scriptsize $\pm$ 0.19 & 47.78 \scriptsize $\pm$ 0.63 & 77.51 \scriptsize $\pm$ 0.28 \\
\cmidrule(lr){1-7}
5 & \checkmark & $\times$ & $\times$ & 1.77 \scriptsize $\pm$ 0.31 & 38.57 \scriptsize $\pm$ 0.58 & 65.48 \scriptsize $\pm$ 1.08 \\
6 & \checkmark & $\times$ & \checkmark & 14.86 \scriptsize $\pm$ 0.36 & \textbf{53.63} \scriptsize $\pm$ 0.16 & \textbf{79.58} \scriptsize $\pm$ 0.23 \\
7 & \checkmark & \checkmark & $\times$ & \underline{15.69} \scriptsize $\pm$ 0.55 & 52.07 \scriptsize $\pm$ 0.40 & 79.18 \scriptsize $\pm$ 0.33 \\
8 & \checkmark & \checkmark & \checkmark & \textbf{15.97} \scriptsize $\pm$ 0.17 & \underline{53.30} \scriptsize $\pm$ 0.29 & \underline{79.56} \scriptsize $\pm$ 0.27 \\

\midrule
\multicolumn{4}{l}{Improvement w.r.t. conf. $1$} & 5.76 \scriptsize $\pm$ 0.18 & 2.81 \scriptsize $\pm$ 0.71 & 2.07 \scriptsize $\pm$ 0.54 \\

\midrule
\multicolumn{7}{c}{\textbf{Horizon} $T=6$} \\
\midrule
1 & $\times$ & $\times$ & $\times$ & 4.70 \scriptsize $\pm$ 0.40 & 45.73 \scriptsize $\pm$ 0.91 & 76.15 \scriptsize $\pm$ 0.33 \\
2 & $\times$ & $\times$ & \checkmark & 4.48 \scriptsize $\pm$ 0.30 & 45.61 \scriptsize $\pm$ 1.10 & 76.02 \scriptsize $\pm$ 0.33 \\
3 & $\times$ & \checkmark & $\times$ & 7.59 \scriptsize $\pm$ 0.32 & 43.66 \scriptsize $\pm$ 0.79 & 76.19 \scriptsize $\pm$ 0.24 \\
4 & $\times$ & \checkmark & \checkmark & 7.20 \scriptsize $\pm$ 0.26 & 43.12 \scriptsize $\pm$ 0.67 & 76.17 \scriptsize $\pm$ 0.22 \\
\cmidrule(lr){1-7}
5 & \checkmark & $\times$ & $\times$ & 0.50 \scriptsize $\pm$ 0.19 & 33.12 \scriptsize $\pm$ 0.75 & 61.08 \scriptsize $\pm$ 1.96 \\
6 & \checkmark & $\times$ & \checkmark & 9.18 \scriptsize $\pm$ 0.26 & \textbf{49.71} \scriptsize $\pm$ 0.36 & \textbf{78.17} \scriptsize $\pm$ 0.17 \\
7 & \checkmark & \checkmark & $\times$ & \underline{9.99} \scriptsize $\pm$ 0.22 & 48.06 \scriptsize $\pm$ 0.37 & 77.58 \scriptsize $\pm$ 0.25 \\
8 & \checkmark & \checkmark & \checkmark & \textbf{10.37} \scriptsize $\pm$ 0.22 & \underline{49.25} \scriptsize $\pm$ 0.54 & \underline{78.01} \scriptsize $\pm$ 0.21 \\

\midrule
\multicolumn{4}{l}{Improvement w.r.t. conf. $1$} & 5.67 \scriptsize $\pm$ 0.43 & 3.52 \scriptsize $\pm$ 1.00 & 1.86 \scriptsize $\pm$ 0.40 \\

\bottomrule
\end{tabular}%
}
\end{table}

\section{Experiments and Results}
\label{appendix:experiments_and_results}

\subsection{Discussion on mIoU metric}
As discussed in MTID~\cite{zhou2025masked}, the definition of mIoU varies across the literature. 
The conventional \emph{set-based} formulation treats each sequence as an unordered set of unique actions:
\begin{equation}
\text{mIoU}_{\text{set}} = \frac{100}{N} \sum_{i=1}^{N} 
\frac{|\tilde{\pi}_i \cap \tilde \pi^{GT}_i|}{|\tilde{\pi}_i \cup \tilde \pi^{GT}_i|},
\end{equation}
where $\tilde{\pi}_i$ and $\tilde \pi^{GT}_i$ denote the predicted and ground-truth actions for sequence $i$, and $N$ is the number of sequences. 
While this formulation captures the overall action coverage, it ignores temporal order and repeated actions. 
This limitation can lead to inflated scores in procedural settings where sequence structure and frequency of actions are crucial (e.g., the action ``\textit{stir mixture}'' may occur multiple times).

To address this limitation, we adopt the \emph{element-wise} (mask-based) IoU formulation
introduced in SCHEMA~\cite{niu2024schema}. In this variant, the IoU is computed
independently for each sequence by comparing the predicted and ground-truth binary masks
along the temporal dimension:
\begin{equation}
\text{mIoU}_{\text{mask}}
=
\frac{100}{N}
\sum_{i=1}^{N}
\frac{
\sum_{t = 1}^{T} \bigl[\tilde{\pi}_{i,t} \land \tilde{\pi}^{GT}_{i,t}\bigr]
}{
\sum_{t = 1}^{T} \bigl[\tilde{\pi}_{i,t} \lor \tilde{\pi}^{GT}_{i,t}\bigr]
+ \varepsilon
},
\label{eq:mask_iou}
\end{equation}
where $\tilde{\pi}_{i, t}$ and $\tilde \pi^{GT}_{i,t}$ denote the predicted and ground-truth action for sequence $i$ at time step \(t\), \(\varepsilon\) is a small constant for numerical stability, and $[\cdot]$ is $1$ if the logical operation between $\tilde{\pi}_{i, t}$ and $\tilde \pi^{GT}_{i,t}$ is true.

Unlike the set-based IoU, this element-wise metric preserves both temporal order and action frequency, offering a more faithful evaluation of sequence prediction, especially in tasks where ordering and repetition are essential.

\subsection{Bootstrap Procedure for Statistical Significance ($fn \ 6$)}
\paragraph{Bootstrap Confidence Intervals for Single-Model Estimates.} To report uncertainty for each model, we compute a bootstrap confidence interval over the scores obtained
from the multiple training seeds. For a given metric (SR, mAcc, or mIoU) and a
fixed planning horizon~$T$, let $\{x_1,\dots,x_n\}$ denote the scores obtained
from $n$ different seeds. The empirical mean is defined as:
\begin{equation}
\bar{x} \;=\; \frac{1}{n}\sum_{i=1}^{n} x_i .
\end{equation}
To estimate the uncertainty around~$\bar{x}$, we perform $K$ bootstrap
resamplings (with $K=10^2$ in all experiments). Each bootstrap replicate is
constructed by sampling with replacement from the original set:
\begin{equation}
X^{*} = \{x^{*}_1,\dots,x^{*}_n\},
\qquad  x^{*}_i \sim \{x_1,\dots,x_n\}.
\end{equation}
For each bootstrap sample, we compute its mean:
\begin{equation}
\mu^{*}_k \;=\; \frac{1}{n}\sum_{i=1}^{n} x^{*}_i,
\qquad k = 1,\dots,K.
\end{equation}
The distribution of the bootstrap means
$\{\mu^{*}_k\}_{k=1}^{K}$ is then used to estimate a
$90\%$ confidence interval by taking the $5$th and $95$th percentiles:
\begin{equation}
\mathrm{CI}_{90}
\;=\;
\bigl[
    \mu^{*}_{(5\%)},
    \;\mu^{*}_{(95\%)}
\bigr].
\end{equation}
In the tables, each metric is reported in the form $\bar{x} \pm \mathrm{CI}$, where $\mathrm{CI} = \mu^{*}_{(95\%)} - \mu^{*}_{(5\%)}$ is the confidence interval width.

\paragraph{Comparison between two models.} To quantify whether the performance differences between two models are statistically significant, we employ a non-parametric bootstrap procedure over the five different training seeds used for each experiment. Let $A = \{a_1,\dots,a_n\}$ and $B = \{b_1,\dots,b_n\}$ denote the seed-level scores (e.g., SR, mAcc, mIoU) for two models, with $n=5$. The observed difference in means is defined as:
\begin{equation}
\Delta_{\mathrm{obs}} \;=\; \frac{1}{n}\sum_{i=1}^{n} a_i \;-\; \frac{1}{n}\sum_{i=1}^{n} b_i.
\end{equation}
To assess its reliability, we generate $K=10^3$ bootstrap replicates. Each replicate samples (with replacement) the sets:
\begin{equation}
A^{*} = \{a^{*}_1,\dots,a^{*}_n\}, \qquad B^{*} = \{b^{*}_1,\dots,b^{*}_n\},
\end{equation}
where $a^{*}_i \sim A$ and $b^{*}_i \sim B$. For each pair of resampled sets, we compute the bootstrap difference
\begin{equation}
\Delta^{*}_k \;=\; \frac{1}{n}\sum_{i=1}^{n} a^{*}_i \;-\; \frac{1}{n}\sum_{i=1}^{n} b^{*}_i, 
\quad k = 1,\dots,K.
\end{equation}
The empirical distribution of $\{\Delta^{*}_k\}_{k=1}^K$ is used to obtain a $90\%$ confidence interval:
\begin{equation}
\mathrm{CI}_{90} \;=\; 
\big[\, \Delta^{*}_{(5\%)} \,,\, \Delta^{*}_{(95\%)} \,\big],
\end{equation}
where $\Delta^{*}_{(p\%)}$ denotes the $p$-th percentile of the bootstrap distribution.  
Improvement in each table is reported in the form $\Delta_{\mathrm{obs}} \pm \mathrm{CI}$. For $\Delta_{\mathrm{obs}}$ each $a_i$ is sampled from our model and each $b_i$ is sampled from the second best model (or from the specified model/configuration). Here, $\mathrm{CI} = \Delta^{*}_{(95\%)} -  \Delta^{*}_{(5\%)}$.
We deem each improvement \emph{statistically significant} if and only if:
\begin{equation}
0 \notin \mathrm{CI}_{90},
\end{equation}
i.e., both endpoints have the same sign.

\subsection{Ablations on CrossTask ($fn \ 8$)}

\paragraph{Importance of Structure-Aware Training.} Table~\ref{tab:ablation_reorg_horizons} reports results for $T \in \{4,5,6\}$ and reveals the same trends observed for $T{=}3$. Across all horizons, three consistent patterns emerge. Together, these findings generalize our conclusions from $T{=}3$ and demonstrate that the benefits of training using DVL persist across longer planning horizons.

\noindent \underline{Structure-aware training is effective.}
Models trained with the Differentiable Viterbi Layer (DVL) (configurations 6-8) outperform their counterparts trained without DVL (configurations 1-4). The absolute gains in SR remain remarkably stable across horizons ($\approx 5.7\%$), indicating that the benefits of structure-aware learning scale reliably with sequence length. In contrast, enabling VD or DVL \emph{only at inference} (configurations 2-4) yields negligible improvements over the baseline, confirming that most of the gains originate from structured training rather than test-time post-processing.

\noindent \underline{DVL learns meaningful emissions.} For every horizon, decoding the learned emissions with a row-wise argmax (configuration 5) leads to a substantial drop in performance, mirroring the behaviour observed for $T{=}3$. This confirms that emissions learned with DVL represent distributions over latent states rather than direct action scores, and therefore require a structured decoding procedure. When these emissions are decoded through VD or DVL (configurations 6-8), performance recovers and consistently exceeds the non-DVL baseline.

\noindent \underline{DVL is Backward-Compatible with standard VD.} Replacing DVL with standard VD at inference (configuration 6 vs.\ 7) results in comparable performance across all metrics and horizons, showing that VD can operate effectively on emissions produced by DVL-trained models. Adding VD on top of DVL (configuration 8) produces only marginal differences. These results confirm that the main advantage stems from structure-aware \emph{training}, with VD and DVL playing largely interchangeable roles at inference.

\begin{figure}[t]
    \centering
    \includegraphics[width=\linewidth]{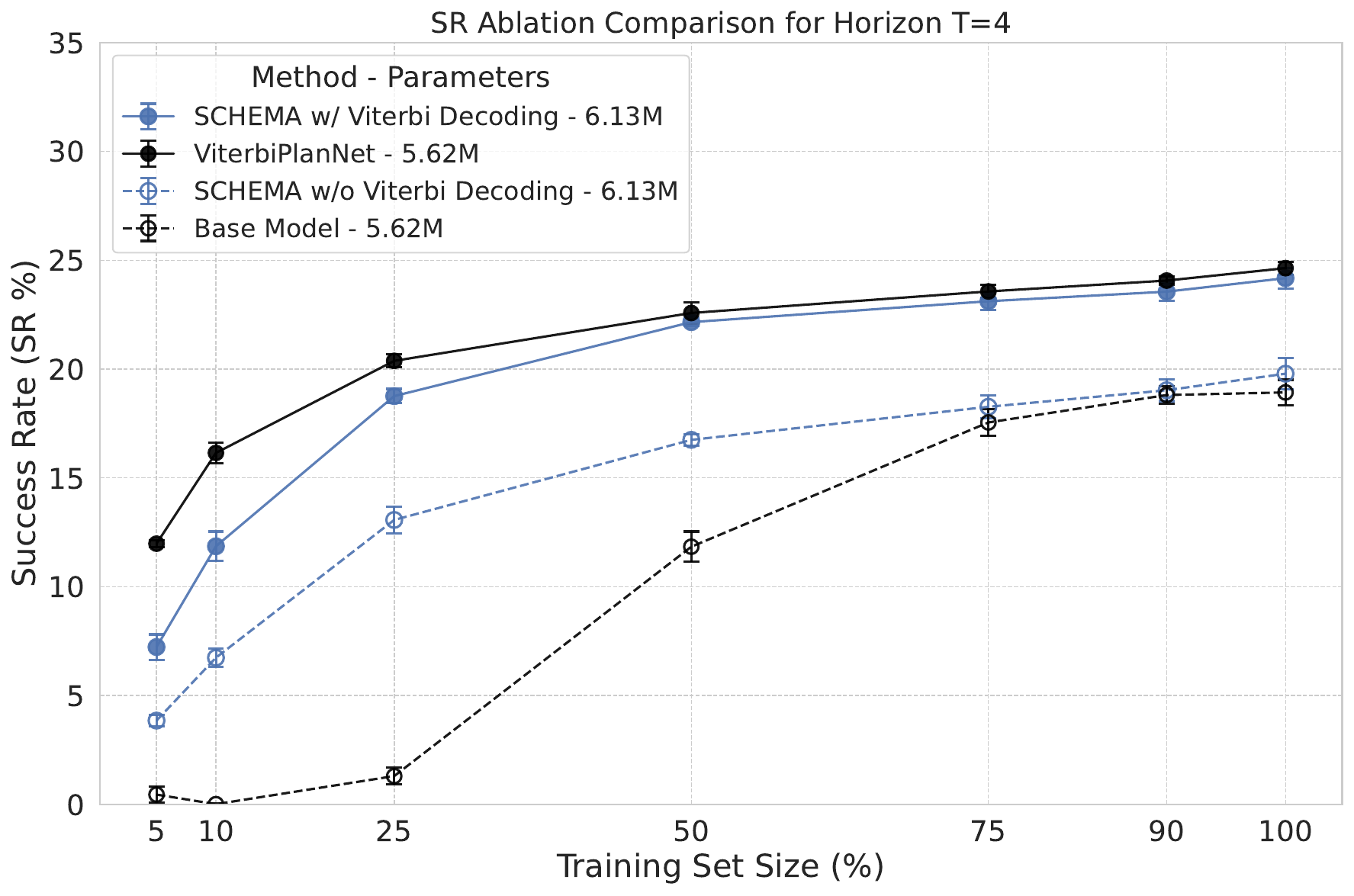}
    \caption{Performance as a function of training data on CrossTask for $T=4$.}
    \label{fig:memorization_T4}
\end{figure}

\begin{figure}[t]
    \centering
    \includegraphics[width=\linewidth]{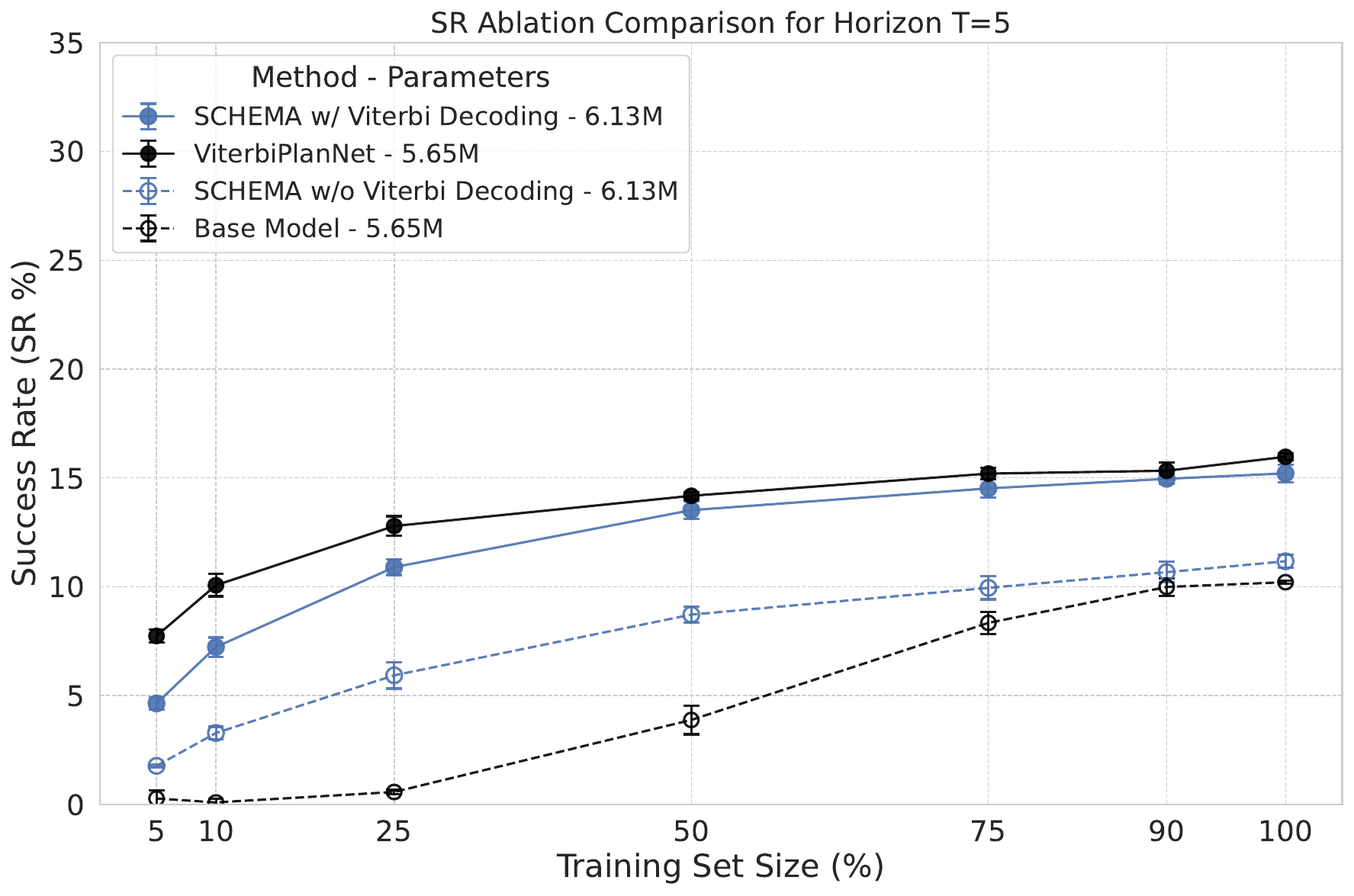}
    \caption{Performance as a function of training data on CrossTask for $T=5$.}
    \label{fig:memorization_T5}
\end{figure}

\begin{figure}[t]
    \centering
    \includegraphics[width=\linewidth]{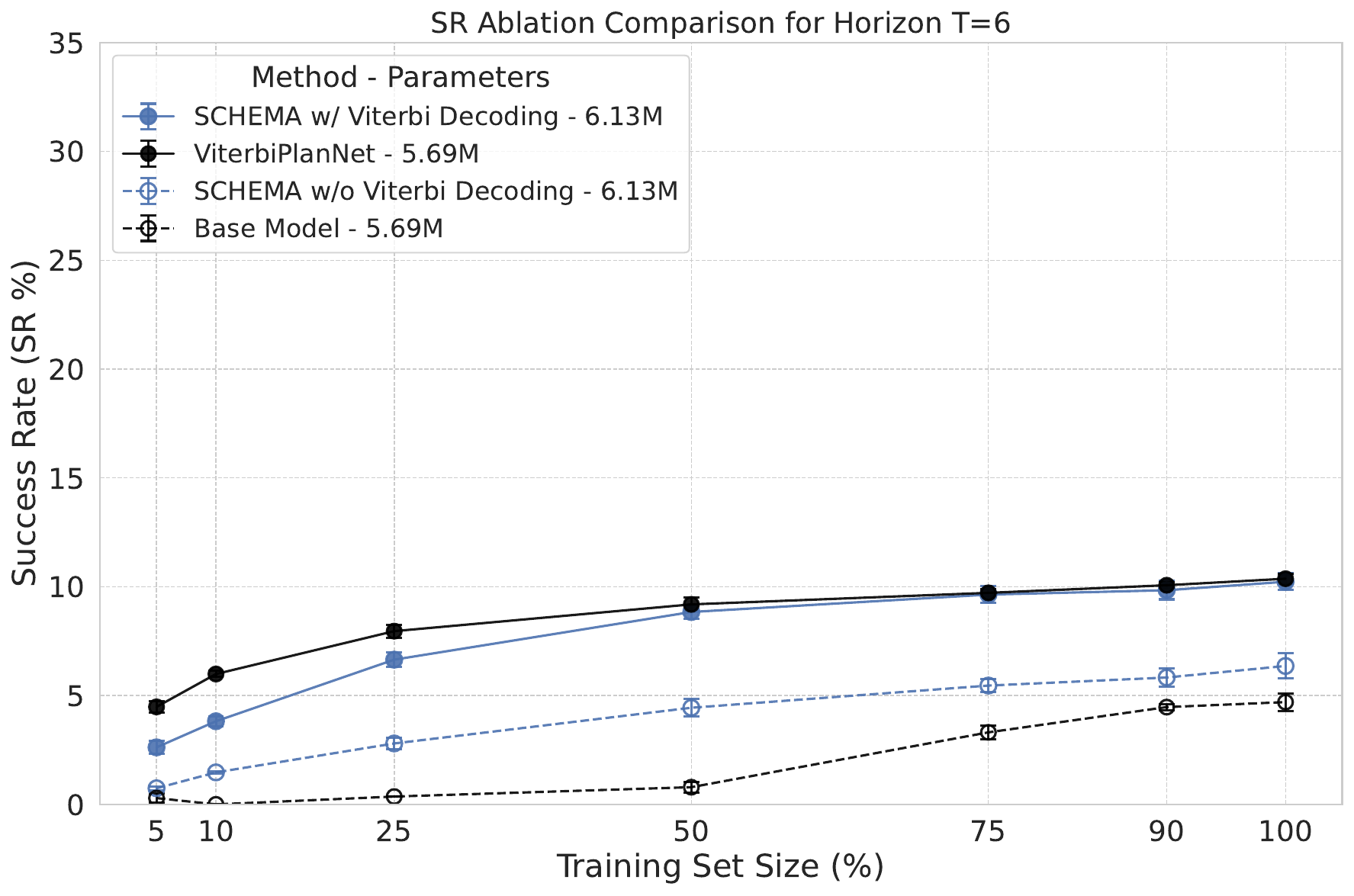}
    \caption{Performance as a function of training data on CrossTask for $T=6$.}
    \label{fig:memorization_T6}
\end{figure}

\paragraph{Memorization and Sample Efficiency.} Figures~\ref{fig:memorization_T4},~\ref{fig:memorization_T5},~\ref{fig:memorization_T6} show performance as a function of training data on CrossTask for $T \in \{4,5,6\}$. They confirm the same patterns observed for $T{=}3$ in the main paper: ViterbiPlanNet is consistently more sample-efficient than SCHEMA. Across all horizons, ViterbiPlanNet achieves higher success rates when trained with limited data (e.g., $5\%\!-\!25\%$ of the training set), while SCHEMA requires substantially more examples to reach comparable performance. This gap progressively narrows as the training set grows, indicating that SCHEMA increasingly benefits from memorization as more trajectories are available.

When the PKG is removed (dashed lines) the results again follow the same trend across all horizons: SCHEMA outperforms the Base Model (corresponding to configuration 1 in Table~\ref{tab:ablation_reorg_horizons}) due to its more flexible transformer-based architecture, which facilitates memorization of procedural patterns. Importantly, the Base Model and ViterbiPlanNet share the same architecture and parameter count, so the consistent improvement of ViterbiPlanNet over the Base Model is entirely attributable to its PKG-aware structured training, rather than additional capacity. These observations, stable across $T{=}4$ and $T{=}5$ as well as $T{=}6$, demonstrate that using the PKG within our differentiable planning module reduces the need for memorization and leads to better sample efficiency across all planning horizons.

\begin{table}[t]
\centering
\caption{SR $\uparrow$ (\%) with and without PKG on CrossTask for Horizons $T \in \{4,5,6\}$.}
\label{tab:sr_comparison_pkg_mult_t}
\resizebox{\linewidth}{!}{%
\begin{tabular}{l c c c c}
\toprule
\textbf{Method} & \textbf{PKG Use} & \textbf{w/o PKG} & \textbf{w/ PKG} & \textbf{Improv.} \\
\midrule
\multicolumn{5}{l}{\textbf{Horizon $T=4$}} \\ %
\midrule
KEPP~\cite{nagasinghe2024not} & Conditioning & 22.57 \scriptsize $\pm$ 0.52 & 22.34 \scriptsize $\pm$ 0.43 & -0.23 \scriptsize $\pm$ 0.69 \\
PlanLLM~\cite{yang2025planllm} & Post-processing & 19.90 \scriptsize $\pm$ 0.40 & 22.91 \scriptsize $\pm$ 1.39 & 3.01 \scriptsize $\pm$ 1.42 \\
SCHEMA~\cite{niu2024schema} & Post-processing & 19.79 \scriptsize $\pm$ 0.72 & 24.18 \scriptsize $\pm$ 0.47 & 4.39 \scriptsize $\pm$ 0.91 \\
\textbf{ViterbiPlanNet} & Guided Training & 18.93 \scriptsize $\pm$ 0.58 & 24.64 \scriptsize $\pm$ 0.30 & 5.71 \scriptsize $\pm$ 0.64 \\
\midrule
\multicolumn{5}{l}{\textbf{Horizon $T=5$}} \\ %
\midrule
KEPP~\cite{nagasinghe2024not} & Conditioning & 13.39 \scriptsize $\pm$ 0.32 & 13.36 \scriptsize $\pm$ 1.06 & -0.03 \scriptsize $\pm$ 1.08 \\
PlanLLM~\cite{yang2025planllm} & Post-processing & 12.06 \scriptsize $\pm$ 0.40 & 14.89 \scriptsize $\pm$ 0.24 & 2.83 \scriptsize $\pm$ 0.45 \\
SCHEMA~\cite{niu2024schema} & Post-processing & 11.17 \scriptsize $\pm$ 0.30 & 15.21 \scriptsize $\pm$ 0.40 & 4.04 \scriptsize $\pm$ 0.51 \\
\textbf{ViterbiPlanNet} & Guided Training & 10.21 \scriptsize $\pm$ 0.08 & 15.97 \scriptsize $\pm$ 0.18 & 5.76 \scriptsize $\pm$ 0.19 \\
\midrule
\multicolumn{5}{l}{\textbf{Horizon $T=6$}} \\ %
\midrule
KEPP~\cite{nagasinghe2024not} & Conditioning & 7.91 \scriptsize $\pm$ 0.66 & 8.21 \scriptsize $\pm$ 0.22 & 0.30 \scriptsize $\pm$ 0.70 \\
PlanLLM~\cite{yang2025planllm} & Post-processing & 7.03 \scriptsize $\pm$ 0.38 & 8.98 \scriptsize $\pm$ 0.97 & 1.95 \scriptsize $\pm$ 1.04 \\
SCHEMA~\cite{niu2024schema} & Post-processing & 6.36 \scriptsize $\pm$ 0.58 & 10.23 \scriptsize $\pm$ 0.38 & 3.87 \scriptsize $\pm$ 0.67 \\
\textbf{ViterbiPlanNet} & Guided Training & 4.70 \scriptsize $\pm$ 0.40 & 10.37 \scriptsize $\pm$ 0.22 & 5.67 \scriptsize $\pm$ 0.43 \\
\bottomrule
\end{tabular}%
}
\end{table}

\paragraph{Guided Training vs Conditioning and Post-processing.} Table~\ref{tab:sr_comparison_pkg_mult_t} extends our comparison of how different methods leverage the PKG to longer horizons ($T \in \{4,5,6\}$), revealing the same pattern observed at $T{=}3$. Across all horizons, we again find that \emph{all} approaches benefit from the PKG, but \emph{not equally}. Methods that use the PKG only as a conditioning signal (KEPP) or as a post-processing constraint (PlanLLM and SCHEMA) exhibit modest gains: KEPP provides almost no improvement, while PlanLLM and SCHEMA obtain consistent but moderate increases. In contrast, ViterbiPlanNet, which incorporates the PKG directly into the training objective through guided training with a Differentiable Viterbi Layer, achieves the \emph{largest} improvement at every horizon. Specifically, ViterbiPlanNet improves by $+5.71\%$, $+5.76\%$, and $+5.67\%$ SR for $T{=}4,5,6$ respectively, substantially surpassing all alternatives.

These results confirm that learning through the PKG is consistently more effective than conditioning on it or applying it only as a test-time constraint. In particular, the advantage of ViterbiPlanNet persists as the planning horizon grows, indicating that guided training extracts a stronger procedural signal and scales more robustly to more challenging horizons.

\begin{table}[t]
\centering
\caption{SR $\uparrow$ (\%) with and without Task Supervision (Task S.) for Horizons $T \in \{3,4,5,6\}$.}
\label{tab:sr_comparison_ts_mult_t}
\begin{tabular}{l l l}
\toprule
\textbf{Method} & \textbf{w/ Task S.} & \textbf{w/o Task S.} \\
\midrule
\multicolumn{3}{l}{\textbf{Horizon $T=3$}} \\
\midrule
PDPP~\cite{wang2023pdpp} & 36.73 \scriptsize $\pm$ 0.59 & 8.39 \scriptsize $\pm$ 8.02 \\
PlanLLM~\cite{yang2025planllm} & 36.84 \scriptsize $\pm$ 1.21 & 15.37 \scriptsize $\pm$ 14.09 \\
SCHEMA~\cite{niu2024schema} & 37.24 \scriptsize $\pm$ 0.60 & 37.00 \scriptsize $\pm$ 0.27 \\
\textbf{ViterbiPlanNet} & 38.45 \scriptsize $\pm$ 0.32 & 38.32 \scriptsize $\pm$ 0.12 \\
\midrule
\multicolumn{3}{l}{\textbf{Horizon $T=4$}} \\
\midrule
PDPP~\cite{wang2023pdpp} & 21.47 \scriptsize $\pm$ 2.09 & 5.24 \scriptsize $\pm$ 1.23 \\
PlanLLM~\cite{yang2025planllm} & 23.25 \scriptsize $\pm$ 0.38 & 4.97 \scriptsize $\pm$ 3.10 \\
SCHEMA~\cite{niu2024schema} & 24.18 \scriptsize $\pm$ 0.47 & 23.24 \scriptsize $\pm$ 0.98 \\
\textbf{ViterbiPlanNet} & 24.64 \scriptsize $\pm$ 0.30 & 24.68 \scriptsize $\pm$ 0.26 \\
\midrule
\multicolumn{3}{l}{\textbf{Horizon $T=5$}} \\
\midrule
PDPP~\cite{wang2023pdpp} & 13.79 \scriptsize $\pm$ 0.21 & 1.51 \scriptsize $\pm$ 1.33 \\
PlanLLM~\cite{yang2025planllm} & 14.88 \scriptsize $\pm$ 0.28 & 5.33 \scriptsize $\pm$ 7.43 \\
SCHEMA~\cite{niu2024schema} & 15.21 \scriptsize $\pm$ 0.40 & 14.80 \scriptsize $\pm$ 0.32 \\
\textbf{ViterbiPlanNet} & 15.97 \scriptsize $\pm$ 0.17 & 15.81 \scriptsize $\pm$ 0.21 \\
\midrule
\multicolumn{3}{l}{\textbf{Horizon $T=6$}} \\
\midrule
PDPP & 8.68 \scriptsize $\pm$ 0.63 & 0.46 \scriptsize $\pm$ 0.22 \\
PlanLLM & 9.23 \scriptsize $\pm$ 0.17 & 2.82 \scriptsize $\pm$ 3.35 \\
SCHEMA & 10.23 \scriptsize $\pm$ 0.38 & 9.32 \scriptsize $\pm$ 0.46 \\
\textbf{ViterbiPlanNet} & 10.37 \scriptsize $\pm$ 0.22 & 10.18 \scriptsize $\pm$ 0.29 \\
\bottomrule
\end{tabular}%
\end{table}

\paragraph{Effect of Task Supervision.} We assess the role of task supervision by %
removing the task head and loss %
during training for all methods. Table~\ref{tab:sr_comparison_ts_mult_t} reports results across multiple planning horizons ($T \in \{3,4,5,6\}$). A clear pattern emerges. Methods whose generation is tightly coupled to the task identity such as PDPP and PlanLLM suffer dramatic performance drops when task labels are removed (e.g., SR at $T{=}3$ for PDPP $36.73\%\!\to\!8.39\%$ and PlanLLM $36.84\%\!\to\!15.37\%$). This happens often with large variance, indicating strong dependence on explicit task conditioning. In contrast, SCHEMA remains remarkably stable across all horizons with changes typically below $1\%$, reflecting the fact that its LLM-derived procedural memory implicitly encodes task-specific structure even without task supervision.

ViterbiPlanNet exhibits the same robustness: performance remains unchanged when task supervision is removed (e.g., $38.45\%\!\to\!38.32\%$ at $T{=}3$, with similar behaviour for larger horizons). This stability stems from the differentiable Viterbi Layer (DVL), which internalizes procedural constraints directly from the PKG and enforces them throughout training. As a result, ViterbiPlanNet learns task-aware structural priors without requiring explicit task annotations.

\begin{table*}[t]
\centering
\caption{Comparison with the state of the art. \textbf{Best} and \underline{second-best} results are highlighted for each metric within each time horizon. Statistically significant performance differences (i.e., cases in which the confidence interval does not include zero) are {\setlength{\fboxsep}{0pt}\colorbox{yellow!30}{marked in yellow}}.}
\label{tab:performance_combined_viterbiplannet_full}
\resizebox{\textwidth}{!}{%
\begin{tabular}{@{}ll lllr lllr lllr@{}}
\toprule
\multirow{2}{*}{\textbf{T}} & \multirow{2}{*}{\textbf{Method}} & \multicolumn{4}{c}{\textbf{CrossTask}} & \multicolumn{4}{c}{\textbf{COIN}} & \multicolumn{4}{c}{\textbf{NIV}} \\
\cmidrule(lr){3-6} \cmidrule(lr){7-10} \cmidrule(lr){11-14}
& & SR $\uparrow$ (\%) & mAcc $\uparrow$ (\%) & mIoU $\uparrow$ (\%) & Params (M) & SR $\uparrow$ (\%) & mAcc $\uparrow$ (\%) & mIoU $\uparrow$ (\%) & Params (M) & SR $\uparrow$ (\%) & mAcc $\uparrow$ (\%) & mIoU $\uparrow$ (\%) & Params (M) \\
\midrule
\multirow{11}{*}{3} 
& Qwen2.5-VL-32B~\cite{bai2025qwen2} & 11.48 & 36.35 & 69.52 & 32,000 & 3.65 & 17.51 & 52.10 & 32,000 & 7.41 & 27.65 & 59.73 & 32,000 \\
& Qwen2.5-32B~\cite{qwen2024qwen2} & 25.14 & 56.10 & 80.92 & 32,000 & 14.97 & 36.34 & 78.74 & 32,000 & 24.07 & 43.46 & 71.88 & 32,000 \\
& Gemini 2.5 Pro~\cite{comanici2025gemini} & 29.18 & 57.90 & 81.48 & $>$100,000 & 17.02 & 38.87 & 78.73 & $>$100,000 & 24.07 & 43.46 & 71.86 & $>$100,000 \\
& Qwen3-30B~\cite{yang2025qwen3} & 23.37 & 55.96 & 81.16 & 30,000 & 14.52 & 36.56 & 78.07 & 30,000 & 24.81 & 42.84 & 70.80 & 30,000 \\
& Qwen3-30B~\cite{yang2025qwen3} + PKG & 23.31 & 56.15 & 81.06 & 30,000 & 14.63 & 36.53 & 78.11 & 30,000 & 25.19 & 43.95 & 71.98 & 30,000 \\
& PKG beam search & 22.38 \scriptsize $\pm$ 0.26 & 55.74 \scriptsize $\pm$ 0.25 & 80.92 \scriptsize $\pm$ 0.26 & 41.87 & 13.32 \scriptsize $\pm$ 0.34 & 37.42 \scriptsize $\pm$ 1.19 & 78.93 \scriptsize $\pm$ 2.06 & 42.90 & 24.96 \scriptsize $\pm$ 1.93 & 43.46 \scriptsize $\pm$ 2.42 & 72.18 \scriptsize $\pm$ 0.55 & 41.74 \\
& PDPP~\cite{wang2023pdpp} & 36.73 \scriptsize $\pm$ 0.59 & 61.96 \scriptsize $\pm$ 0.59 & 83.20 \scriptsize $\pm$ 0.33 & 41.87 & 22.37 \scriptsize $\pm$ 0.57 & 44.60 \scriptsize $\pm$ 0.16 & 83.00 \scriptsize $\pm$ 0.42 & 42.90 & 26.52 \scriptsize $\pm$ 1.56 & 45.58 \scriptsize $\pm$ 1.85 & \textbf{74.89} \scriptsize $\pm$ 0.85 & 41.74 \\
& KEPP~\cite{nagasinghe2024not} & 34.93 \scriptsize $\pm$ 2.60 & 60.34 \scriptsize $\pm$ 1.61 & 82.67 \scriptsize $\pm$ 0.69 & 42.18 & 13.85 \scriptsize $\pm$ 7.49 & 28.40 \scriptsize $\pm$ 12.26 & 62.54 \scriptsize $\pm$ 14.35 & 44.66 & 27.56 \scriptsize $\pm$ 1.48 & \underline{45.93} \scriptsize $\pm$ 2.37 & \underline{74.36} \scriptsize $\pm$ 0.97 & 41.86 \\
& PlanLLM~\cite{yang2025planllm} & 36.84 \scriptsize $\pm$ 1.21 & 61.56 \scriptsize $\pm$ 1.03 & 83.23 \scriptsize $\pm$ 0.53 & 384.94 & \underline{33.44} \scriptsize $\pm$ 0.15 & \textbf{51.05} \scriptsize $\pm$ 0.46 & \textbf{84.66} \scriptsize $\pm$ 0.41 & 386.43 & \underline{30.00} \scriptsize $\pm$ 1.41 & 44.35 \scriptsize $\pm$ 2.52 & 73.60 \scriptsize $\pm$ 1.66 & 384.77 \\
& SCHEMA~\cite{niu2024schema} & \underline{37.24} \scriptsize $\pm$ 0.60 & \underline{62.69} \scriptsize $\pm$ 0.28 & \textbf{83.94} \scriptsize $\pm$ 0.23 & 6.13 & 32.89 \scriptsize $\pm$ 0.61 & 50.84 \scriptsize $\pm$ 0.47 & \underline{83.98} \scriptsize $\pm$ 0.67 & 6.28 & 26.30 \scriptsize $\pm$ 1.49 & 42.77 \scriptsize $\pm$ 2.12 & 73.04 \scriptsize $\pm$ 1.42 & 6.12 \\
\highlightrow
& ViterbiPlanNet & \textbf{38.45} \scriptsize $\pm$ 0.32 & \textbf{63.07} \scriptsize $\pm$ 0.17 & \underline{83.89} \scriptsize $\pm$ 0.16 & 5.57 & \textbf{33.99} \scriptsize $\pm$ 0.23 & \underline{50.87} \scriptsize $\pm$ 0.17 & 83.88 \scriptsize $\pm$ 0.31 & 6.67 & \textbf{32.37} \scriptsize $\pm$ 0.96 & \textbf{46.96} \scriptsize $\pm$ 1.75 & 73.85 \scriptsize $\pm$ 0.85 & 5.48 \\
\rowcolor{gray!10}
& Improvement & \colorbox{yellow!30}{+1.21 \scriptsize $\pm$ 0.69} & \colorbox{yellow!30}{+0.38 \scriptsize $\pm$ 0.34} & -0.05 \scriptsize $\pm$ 0.27 &  & \colorbox{yellow!30}{+0.55 \scriptsize $\pm$ 0.27} & -0.18 \scriptsize $\pm$ 0.49 & \colorbox{yellow!30}{-0.78 \scriptsize $\pm$ 0.50} &  & \colorbox{yellow!30}{+2.37 \scriptsize $\pm$ 1.63} & +1.04 \scriptsize $\pm$ 3.06 & \colorbox{yellow!30}{-1.04 \scriptsize $\pm$ 1.22} &  \\
\midrule
\multirow{11}{*}{4} 
& Qwen2.5-VL-32B~\cite{bai2025qwen2} & 5.56 & 31.22 & 66.31 & 32,000 & 1.87 & 17.05 & 55.66 & 32,000 & 5.26 & 28.84 & 60.21 & 32,000 \\
& Qwen2.5-32B~\cite{qwen2024qwen2} & 9.22 & 46.32 & 76.15 & 32,000 & 4.98 & 27.45 & 71.64 & 32,000 & 23.25 & 41.89 & 73.91 & 32,000 \\
& Gemini 2.5 Pro~\cite{comanici2025gemini} & 14.00 & 51.33 & 78.58 & $>$100,000 & 8.10 & 31.90 & 71.70 & $>$100,000 & 22.37 & 40.35 & 73.05 & $>$100,000 \\
& Qwen3-30B~\cite{yang2025qwen3} & 10.59 & 49.06 & 78.03 & 30,000 & 4.64 & 28.85 & 70.45 & 30,000 & 22.37 & 41.23 & 73.90 & 30,000 \\
& Qwen3-30B~\cite{yang2025qwen3} + PKG & 10.96 & 48.77 & 77.48 & 30,000 & 4.78 & 29.00 & 71.04 & 30,000 & 21.93 & 41.67 & 74.43 & 30,000 \\
& PKG beam search & 9.30 \scriptsize $\pm$ 0.22 & 47.65 \scriptsize $\pm$ 0.54 & 78.25 \scriptsize $\pm$ 0.42 & 41.87 & 5.14 \scriptsize $\pm$ 0.60 & 31.29 \scriptsize $\pm$ 3.64 & 74.26 \scriptsize $\pm$ 5.38 & 42.90 & 21.23 \scriptsize $\pm$ 0.96 & 40.86 \scriptsize $\pm$ 0.83 & 72.69 \scriptsize $\pm$ 0.75 & 41.74 \\
& PDPP~\cite{wang2023pdpp} & 21.47 \scriptsize $\pm$ 2.09 & 55.66 \scriptsize $\pm$ 1.64 & 80.68 \scriptsize $\pm$ 0.83 & 41.87 & 15.21 \scriptsize $\pm$ 0.34 & 41.01 \scriptsize $\pm$ 0.32 & 81.64 \scriptsize $\pm$ 0.48 & 42.90 & 21.40 \scriptsize $\pm$ 0.53 & 40.20 \scriptsize $\pm$ 2.00 & 72.82 \scriptsize $\pm$ 1.84 & 41.74 \\
& KEPP~\cite{nagasinghe2024not} & 22.34 \scriptsize $\pm$ 0.43 & 55.24 \scriptsize $\pm$ 0.30 & 80.58 \scriptsize $\pm$ 0.25 & 42.18 & 15.20 \scriptsize $\pm$ 1.27 & 33.39 \scriptsize $\pm$ 0.73 & 67.79 \scriptsize $\pm$ 1.29 & 44.66 & 22.54 \scriptsize $\pm$ 1.93 & \underline{42.46} \scriptsize $\pm$ 1.49 & 73.11 \scriptsize $\pm$ 0.94 & 41.86 \\
& PlanLLM~\cite{yang2025planllm} & 22.91 \scriptsize $\pm$ 1.39 & 55.29 \scriptsize $\pm$ 1.54 & 81.03 \scriptsize $\pm$ 0.47 & 384.94 & \underline{23.19} \scriptsize $\pm$ 0.32 & \textbf{45.70} \scriptsize $\pm$ 0.33 & \textbf{83.44} \scriptsize $\pm$ 0.39 & 386.43 & 23.42 \scriptsize $\pm$ 1.40 & 41.95 \scriptsize $\pm$ 2.81 & 72.32 \scriptsize $\pm$ 0.91 & 384.77 \\
& SCHEMA~\cite{niu2024schema} & \underline{24.18} \scriptsize $\pm$ 0.47 & \textbf{57.02} \scriptsize $\pm$ 0.64 & \textbf{81.46} \scriptsize $\pm$ 0.19 & 6.13 & 22.33 \scriptsize $\pm$ 0.92 & 45.21 \scriptsize $\pm$ 1.05 & \underline{82.93} \scriptsize $\pm$ 0.25 & 6.28 & \underline{24.39} \scriptsize $\pm$ 1.84 & 41.14 \scriptsize $\pm$ 3.62 & \underline{73.13} \scriptsize $\pm$ 1.97 & 6.12 \\
\highlightrow
& ViterbiPlanNet & \textbf{24.64} \scriptsize $\pm$ 0.30 & \underline{57.00} \scriptsize $\pm$ 0.42 & \underline{81.18} \scriptsize $\pm$ 0.44 & 5.60 & \textbf{23.92} \scriptsize $\pm$ 0.29 & \underline{45.63} \scriptsize $\pm$ 0.55 & 82.56 \scriptsize $\pm$ 0.44 & 6.87 & \textbf{27.54} \scriptsize $\pm$ 0.70 & \textbf{45.55} \scriptsize $\pm$ 1.89 & \textbf{74.71} \scriptsize $\pm$ 1.19 & 5.50 \\
\rowcolor{gray!10}
& Improvement & +0.46 \scriptsize $\pm$ 0.61 & -0.02 \scriptsize $\pm$ 0.78 & -0.29 \scriptsize $\pm$ 0.49 &  & \colorbox{yellow!30}{+0.73 \scriptsize $\pm$ 0.44} & -0.08 \scriptsize $\pm$ 0.62 & \colorbox{yellow!30}{-0.88 \scriptsize $\pm$ 0.59} &  & \colorbox{yellow!30}{+3.15 \scriptsize $\pm$ 1.93} & \colorbox{yellow!30}{+3.09 \scriptsize $\pm$ 2.43} & +1.58 \scriptsize $\pm$ 2.37 &  \\
\midrule
\multirow{11}{*}{5} 
& Qwen2.5-VL-32B~\cite{bai2025qwen2} & 2.27 & 24.53 & 63.65 & 32,000 & 1.10 & 16.75 & 57.21 & 32,000 & 1.07 & 29.30 & 61.98 & 32,000 \\
& Qwen2.5-32B~\cite{qwen2024qwen2} & 3.77 & 37.58 & 72.09 & 32,000 & 4.04 & 28.48 & 74.77 & 32,000 & 18.72 & 43.32 & 73.47 & 32,000 \\
& Gemini 2.5 Pro~\cite{comanici2025gemini} & 4.35 & 42.76 & 76.47 & $>$100,000 & 7.89 & 32.48 & 75.22 & $>$100,000 & 18.72 & 40.86 & 71.71 & $>$100,000 \\
& Qwen3-30B~\cite{yang2025qwen3} & 4.09 & 39.14 & 72.83 & 30,000 & 3.65 & 28.46 & 72.44 & 30,000 & 19.25 & 43.21 & 72.90 & 30,000 \\
& Qwen3-30B~\cite{yang2025qwen3} + PKG & 4.77 & 39.83 & 73.17 & 30,000 & 3.55 & 29.08 & 74.06 & 30,000 & 20.32 & 45.35 & 74.25 & 30,000 \\
& PKG beam search & 5.35 \scriptsize $\pm$ 0.03 & 42.98 \scriptsize $\pm$ 0.69 & 76.75 \scriptsize $\pm$ 0.54 & 41.87 & 2.91 \scriptsize $\pm$ 0.19 & 28.93 \scriptsize $\pm$ 0.93 & 73.61 \scriptsize $\pm$ 1.92 & 42.90 & 18.40 \scriptsize $\pm$ 1.18 & 42.65 \scriptsize $\pm$ 1.48 & 74.00 \scriptsize $\pm$ 1.58 & 41.74 \\
& PDPP~\cite{wang2023pdpp} & 13.79 \scriptsize $\pm$ 0.21 & 52.31 \scriptsize $\pm$ 0.29 & 79.21 \scriptsize $\pm$ 0.26 & 41.87 & 11.42 \scriptsize $\pm$ 0.49 & 37.23 \scriptsize $\pm$ 0.34 & 80.84 \scriptsize $\pm$ 0.47 & 42.90 & 19.04 \scriptsize $\pm$ 2.57 & \textbf{44.56} \scriptsize $\pm$ 2.97 & \textbf{75.73} \scriptsize $\pm$ 1.35 & 41.74 \\
& KEPP~\cite{nagasinghe2024not} & 13.36 \scriptsize $\pm$ 1.06 & 51.27 \scriptsize $\pm$ 0.73 & 78.69 \scriptsize $\pm$ 0.62 & 42.18 & 12.14 \scriptsize $\pm$ 0.35 & 32.28 \scriptsize $\pm$ 0.57 & 69.19 \scriptsize $\pm$ 0.94 & 44.66 & 21.07 \scriptsize $\pm$ 1.39 & \underline{44.36} \scriptsize $\pm$ 2.25 & \underline{74.93} \scriptsize $\pm$ 1.42 & 41.86 \\
& PlanLLM~\cite{yang2025planllm} & 14.89 \scriptsize $\pm$ 0.24 & 51.16 \scriptsize $\pm$ 0.70 & 78.97 \scriptsize $\pm$ 0.35 & 384.94 & \textbf{16.15} \scriptsize $\pm$ 0.41 & \textbf{40.29} \scriptsize $\pm$ 0.86 & \textbf{82.21} \scriptsize $\pm$ 1.67 & 386.43 & \underline{21.93} \scriptsize $\pm$ 0.43 & 42.89 \scriptsize $\pm$ 1.22 & 73.84 \scriptsize $\pm$ 1.51 & 384.77 \\
& SCHEMA~\cite{niu2024schema} & \underline{15.21} \scriptsize $\pm$ 0.40 & \underline{52.97} \scriptsize $\pm$ 0.24 & \underline{79.44} \scriptsize $\pm$ 0.29 & 6.13 & 15.30 \scriptsize $\pm$ 0.73 & \underline{39.47} \scriptsize $\pm$ 0.89 & \underline{81.27} \scriptsize $\pm$ 1.09 & 6.28 & 19.14 \scriptsize $\pm$ 0.97 & 39.25 \scriptsize $\pm$ 3.34 & 72.97 \scriptsize $\pm$ 2.34 & 6.12 \\
\highlightrow
& ViterbiPlanNet & \textbf{15.97} \scriptsize $\pm$ 0.17 & \textbf{53.30} \scriptsize $\pm$ 0.29 & \textbf{79.56} \scriptsize $\pm$ 0.27 & 5.64 & \underline{15.87} \scriptsize $\pm$ 0.53 & 39.42 \scriptsize $\pm$ 0.69 & 81.19 \scriptsize $\pm$ 0.99 & 7.07 & \textbf{23.10} \scriptsize $\pm$ 0.64 & 42.97 \scriptsize $\pm$ 1.99 & 74.81 \scriptsize $\pm$ 1.28 & 5.51 \\
\rowcolor{gray!10}
& Improvement & \colorbox{yellow!30}{+0.76 \scriptsize $\pm$ 0.43} & +0.33 \scriptsize $\pm$ 0.40 & +0.12 \scriptsize $\pm$ 0.40 &  & -0.21 \scriptsize $\pm$ 0.70 & -0.93 \scriptsize $\pm$ 1.29 & -1.23 \scriptsize $\pm$ 2.04 &  & \colorbox{yellow!30}{+1.18 \scriptsize $\pm$ 0.86} & -1.59 \scriptsize $\pm$ 3.47 & -0.92 \scriptsize $\pm$ 1.82 &  \\
\midrule
\multirow{11}{*}{6} 
& Qwen2.5-VL-32B~\cite{bai2025qwen2} & 1.21 & 25.17 & 63.01 & 32,000 & 0.57 & 16.46 & 57.35 & 32,000 & 3.38 & 32.09 & 65.33 & 32,000 \\
& Qwen2.5-32B~\cite{qwen2024qwen2} & 4.00 & 38.71 & 73.50 & 32,000 & 5.12 & 27.27 & 69.84 & 32,000 & 16.22 & 42.23 & 72.29 & 32,000 \\
& Gemini 2.5 Pro~\cite{comanici2025gemini} & 3.84 & 38.55 & 74.54 & $>$100,000 & 8.79 & 30.75 & 70.24 & $>$100,000 & 16.89 & 40.20 & 70.31 & $>$100,000 \\
& Qwen3-30B~\cite{yang2025qwen3} & 3.40 & 39.09 & 74.31 & 30,000 & 1.91 & 25.97 & 68.30 & 30,000 & 16.89 & 41.89 & 71.28 & 30,000 \\
& Qwen3-30B~\cite{yang2025qwen3} + PKG & 3.52 & 39.52 & 74.46 & 30,000 & 2.67 & 26.52 & 68.47 & 30,000 & 17.57 & 43.02 & 72.20 & 30,000 \\
& PKG beam search & 2.65 \scriptsize $\pm$ 0.16 & 38.64 \scriptsize $\pm$ 0.52 & 74.89 \scriptsize $\pm$ 0.54 & 41.87 & 0.69 \scriptsize $\pm$ 0.18 & 25.99 \scriptsize $\pm$ 0.11 & 71.25 \scriptsize $\pm$ 1.31 & 42.90 & 15.41 \scriptsize $\pm$ 0.41 & 42.23 \scriptsize $\pm$ 2.68 & 72.26 \scriptsize $\pm$ 1.13 & 41.74 \\
& PDPP~\cite{wang2023pdpp} & 8.68 \scriptsize $\pm$ 0.63 & 48.50 \scriptsize $\pm$ 1.15 & \underline{78.10} \scriptsize $\pm$ 0.59 & 41.87 & 9.14 \scriptsize $\pm$ 0.44 & 33.83 \scriptsize $\pm$ 0.54 & 78.42 \scriptsize $\pm$ 0.26 & 42.90 & 14.19 \scriptsize $\pm$ 1.76 & \underline{43.83} \scriptsize $\pm$ 2.43 & \underline{74.31} \scriptsize $\pm$ 1.56 & 41.74 \\
& KEPP~\cite{nagasinghe2024not} & 8.21 \scriptsize $\pm$ 0.22 & 46.45 \scriptsize $\pm$ 1.11 & 76.45 \scriptsize $\pm$ 0.74 & 42.18 & 10.16 \scriptsize $\pm$ 0.53 & 30.99 \scriptsize $\pm$ 0.95 & 69.40 \scriptsize $\pm$ 1.28 & 44.66 & 14.05 \scriptsize $\pm$ 0.81 & 41.24 \scriptsize $\pm$ 1.49 & 73.44 \scriptsize $\pm$ 1.04 & 41.86 \\
& PlanLLM~\cite{yang2025planllm} & 9.04 \scriptsize $\pm$ 0.82 & 45.91 \scriptsize $\pm$ 1.24 & 76.91 \scriptsize $\pm$ 0.65 & 384.94 & 12.51 \scriptsize $\pm$ 0.24 & 34.97 \scriptsize $\pm$ 0.42 & 78.17 \scriptsize $\pm$ 0.48 & 386.43 & \underline{16.35} \scriptsize $\pm$ 0.81 & 40.79 \scriptsize $\pm$ 1.26 & 73.52 \scriptsize $\pm$ 0.97 & 384.77 \\
& SCHEMA~\cite{niu2024schema} & \underline{10.23} \scriptsize $\pm$ 0.38 & \textbf{49.31} \scriptsize $\pm$ 0.49 & \textbf{78.31} \scriptsize $\pm$ 0.34 & 6.13 & \textbf{13.16} \scriptsize $\pm$ 0.49 & \textbf{36.41} \scriptsize $\pm$ 0.60 & \underline{79.20} \scriptsize $\pm$ 0.85 & 6.28 & 15.81 \scriptsize $\pm$ 2.97 & 40.20 \scriptsize $\pm$ 3.99 & 73.46 \scriptsize $\pm$ 1.49 & 6.12 \\
\highlightrow
& ViterbiPlanNet & \textbf{10.37} \scriptsize $\pm$ 0.22 & \underline{49.25} \scriptsize $\pm$ 0.54 & 78.01 \scriptsize $\pm$ 0.21 & 5.67 & \underline{13.11} \scriptsize $\pm$ 0.35 & \underline{36.03} \scriptsize $\pm$ 0.39 & \textbf{79.35} \scriptsize $\pm$ 0.64 & 7.27 & \textbf{18.78} \scriptsize $\pm$ 0.81 & \textbf{45.77} \scriptsize $\pm$ 0.83 & \textbf{75.91} \scriptsize $\pm$ 0.52 & 5.52 \\
\rowcolor{gray!10}
& Improvement & +0.14 \scriptsize $\pm$ 0.44 & -0.06 \scriptsize $\pm$ 0.73 & -0.30 \scriptsize $\pm$ 0.41 &  & -0.05 \scriptsize $\pm$ 0.57 & -0.38 \scriptsize $\pm$ 0.73 & +0.15 \scriptsize $\pm$ 1.09 &  & \colorbox{yellow!30}{+2.43 \scriptsize $\pm$ 1.08} & +1.94 \scriptsize $\pm$ 2.55 & +1.60 \scriptsize $\pm$ 1.66 &  \\
\bottomrule
\end{tabular}%
}
\end{table*}

\begin{table*}[t!]
\centering
\caption{Performance comparison of MTID$^{\clubsuit}$ and ViterbiPlanNet$^{\clubsuit}$ across the CrossTask, COIN, and NIV datasets. The \textbf{best} and \underline{second-best} results are highlighted for each metric within each time horizon.}
\label{tab:performance_comparison_MTID_full}
\resizebox{\textwidth}{!}{%
\begin{tabular}{@{}ll cccr cccr cccr@{}}
\toprule
\multirow{2}{*}{\textbf{Horizon}} & \multirow{2}{*}{\textbf{Method}} & \multicolumn{4}{c}{\textbf{CrossTask}} & \multicolumn{4}{c}{\textbf{COIN}} & \multicolumn{4}{c}{\textbf{NIV}} \\
\cmidrule(lr){3-6} \cmidrule(lr){7-10} \cmidrule(lr){11-14}
& & SR $\uparrow$ (\%) & mAcc $\uparrow$ (\%) & mIoU $\uparrow$ (\%) & Params (M) & SR $\uparrow$ (\%) & mAcc $\uparrow$ (\%) & mIoU $\uparrow$ (\%) & Params (M) & SR $\uparrow$ (\%) & mAcc $\uparrow$ (\%) & mIoU $\uparrow$ (\%) & Params (M) \\
\midrule
\multirow{2}{*}{T = 3} & MTID$^{\clubsuit}$~\cite{zhou2025masked} & \textbf{40.45} & \underline{67.19} & \underline{69.17} & 1085.20 & \underline{30.44} & \textbf{51.70} & \underline{59.74} & 1085.20 & \underline{28.52} & \underline{44.44} & \underline{56.46} & 1085.20 \\
& ViterbiPlanNet$^{\clubsuit}$ & \underline{39.75} & \textbf{67.39} & \textbf{76.92} & 5.49 & \textbf{34.42} & \underline{51.20} & \textbf{81.25} & 6.01 & \textbf{34.44} & \textbf{48.89} & \textbf{95.07} & 5.45 \\
\midrule
\multirow{2}{*}{T = 4} & MTID$^{\clubsuit}$~\cite{zhou2025masked} & \textbf{24.76} & \underline{60.69} & \underline{67.67} & 1085.20 & \underline{22.74} & \textbf{49.90} & \underline{61.25} & 1085.20 & \underline{24.89} & \underline{44.54} & \underline{57.46} & 1085.20 \\
& ViterbiPlanNet$^{\clubsuit}$ & \underline{24.19} & \textbf{61.12} & \textbf{80.67} & 5.52 & \textbf{24.09} & \underline{45.71} & \textbf{77.84} & 6.21 & \textbf{28.95} & \textbf{47.81} & \textbf{80.07} & 5.46 \\
\midrule
\multirow{2}{*}{T = 5} & MTID$^{\clubsuit}$~\cite{zhou2025masked} & \underline{15.26} & - & - & 1085.20 & \multicolumn{4}{c}{-} & \multicolumn{4}{c}{-} \\
& ViterbiPlanNet$^{\clubsuit}$ & \textbf{15.37} & 57.10 & 80.53 & 5.55 & \multicolumn{4}{c}{-} & \multicolumn{4}{c}{-} \\
\midrule
\multirow{2}{*}{T = 6} & MTID$^{\clubsuit}$~\cite{zhou2025masked} & \textbf{10.30} & - & - & 1085.20 & \multicolumn{4}{c}{-} & \multicolumn{4}{c}{-} \\
& ViterbiPlanNet$^{\clubsuit}$ & \underline{9.68} & 53.99 & 86.34 & 5.57 & \multicolumn{4}{c}{-} & \multicolumn{4}{c}{-} \\
\bottomrule
\end{tabular}%
}
\end{table*}

\subsection{Comparisons with the State of the Art ($fn\ 8$)}
\paragraph{Performance on Different Planning Horizons.} Table~\ref{tab:performance_combined_viterbiplannet_full} reports the full comparison across all datasets for horizons beyond those shown in the main paper. The trends observed for $T{=}3$ and $T{=}4$ remain fully consistent at larger horizons.

\noindent \underline{ViterbiPlanNet achieves the highest Success Rate (SR)} across all settings. Although performance gaps naturally shrink as the task becomes harder and all methods degrade, ViterbiPlanNet continues to match or surpass the strongest baselines, typically SCHEMA or PlanLLM, and remains the top performer in SR, demonstrating robust long-horizon sequential modeling.

\noindent \underline{Step-level metrics (mAcc and mIoU)} remain comparable to other approaches. Similar to shorter horizons, SCHEMA and PlanLLM occasionally report slightly higher mAcc or mIoU on COIN, but differences are small and not statistically significant. This confirms that ViterbiPlanNet's emphasis on global procedural consistency does not compromise local accuracy, even for longer sequences.

\noindent \underline{In-context LLM/VLM models achieve limited performance.} At $T{=}5$ and especially $T{=}6$, Qwen2.5-VL-32B, Qwen LLMs, and Gemini~2.5~Pro exhibit large drops in SR, often falling below the simple PKG beam-search baseline. This highlights that zero-shot prompting strategies struggle to maintain coherent multi-step reasoning as planning depth increases.

\begin{table}[t]
\centering
\caption{Cross-Horizon Consistency results on COIN.}
\label{tab:horizon_comparison_coin}
\resizebox{\linewidth}{!}{%
\begin{tabular}{l l l l}
\toprule
\textbf{Method} & \textbf{SR} $\uparrow$ \textbf{(\%) [6 $\rightarrow$ 3]} & \textbf{SR} $\uparrow$ \textbf{(\%) [6 $\rightarrow$ 4]} & \textbf{SR} $\uparrow$ \textbf{(\%) [6 $\rightarrow$ 5]} \\
\midrule
Qwen2.5-VL-32B~\cite{bai2025qwen2} & 3.47 & 1.75 & 1.00 \\
Qwen2.5-32B~\cite{qwen2024qwen2} & 6.71 & 6.55 & 3.75 \\
Gemini 2.5 Pro~\cite{comanici2025gemini} & 5.35 & 7.20 & 2.12 \\
Qwen3-30B~\cite{yang2025qwen3} & 5.82 & 4.31 & 2.51 \\
Qwen3-30B~\cite{yang2025qwen3} + PKG & 5.80 & 4.84 & 2.71 \\
PKG beam search & 6.85 \scriptsize $\pm$ 0.27 & 4.39 \scriptsize $\pm$ 0.14 & 1.66 \scriptsize $\pm$ 0.20 \\
\midrule
PDPP~\cite{wang2023pdpp} & 7.66 \scriptsize $\pm$ 0.17 & 6.84 \scriptsize $\pm$ 0.36 & 5.00 \scriptsize $\pm$ 0.49 \\
KEPP~\cite{nagasinghe2024not} & 5.10 \scriptsize $\pm$ 0.80 & 6.09 \scriptsize $\pm$ 1.04 & 5.85 \scriptsize $\pm$ 1.03 \\
PlanLLM~\cite{yang2025planllm} & 9.48 \scriptsize $\pm$ 0.38 & 8.22 \scriptsize $\pm$ 0.27 & 6.02 \scriptsize $\pm$ 0.51 \\
SCHEMA~\cite{niu2024schema} & \underline{9.89} \scriptsize $\pm$ 0.82 & \underline{9.30} \scriptsize $\pm$ 1.14 & \underline{7.89} \scriptsize $\pm$ 0.64 \\
\textbf{ViterbiPlanNet} & \textbf{14.34} \scriptsize $\pm$ 0.41 & \textbf{13.82} \scriptsize $\pm$ 0.53 & \textbf{9.47} \scriptsize $\pm$ 0.14 \\
\midrule
Improvement & +4.45 \scriptsize $\pm$ 0.90 & +4.52 \scriptsize $\pm$ 1.31 & +1.58 \scriptsize $\pm$ 0.67 \\
\bottomrule
\end{tabular}%
}
\end{table}

\paragraph{Cross-Horizon Consistency.} Tables~\ref{tab:horizon_comparison_coin} and~\ref{tab:horizon_comparison_niv} report cross-horizon consistency results on COIN and NIV, extending the analysis from the main paper. The same trends observed on CrossTask clearly emerge across both datasets. On COIN, LLMs and VLMs exhibit limited robustness when evaluated at shorter horizons after training at $T{=}6$, with performance often collapsing as the required subsequence length decreases. Learning-based approaches such as PDPP, KEPP, and PlanLLM show slightly better stability but still suffer from noticeable degradation, particularly when moving from $6{\rightarrow}3$. SCHEMA stands out as the strongest baseline, yet ViterbiPlanNet consistently surpasses it by substantial margins across all horizon reductions, achieving gains of up to $+4.44\%$ SR. 

On NIV, the role of explicit procedural structure is particularly evident. PKG beam search stands out as the strongest non-learning baseline and, in fact, the second-best overall method, clearly demonstrating the high importance of the PKG signal on this dataset. This indicates that NIV strongly benefits from structured graph-based priors. Importantly, ViterbiPlanNet is the only model that fully leverages this signal. The improvement is statistically significant for the $6{\rightarrow}3$ setting ($+3.85\%$ SR), while for $6{\rightarrow}4$ and $6{\rightarrow}5$ the gains remain positive but not statistically conclusive. These results underscore that ViterbiPlanNet is uniquely capable of leveraging the PKG to achieve robust horizon-invariant planning behavior.

\begin{table}[t]
\centering
\caption{Cross-Horizon Consistency results on NIV.}
\label{tab:horizon_comparison_niv}
\resizebox{\linewidth}{!}{%
\begin{tabular}{l l l l}
\toprule
\textbf{Method} & \textbf{SR} $\uparrow$ \textbf{(\%) [6 $\rightarrow$ 3]} & \textbf{SR} $\uparrow$ \textbf{(\%) [6 $\rightarrow$ 4]} & \textbf{SR} $\uparrow$ \textbf{(\%) [6 $\rightarrow$ 5]} \\
\midrule
Qwen2.5-VL-32B~\cite{bai2025qwen2} & 9.26 & 5.70 & 1.60 \\
Qwen2.5-32B~\cite{qwen2024qwen2} & 3.33 & 5.70 & 8.02 \\
Gemini 2.5 Pro~\cite{comanici2025gemini} & 1.11 & 6.14 & 9.63 \\
Qwen3-30B~\cite{yang2025qwen3} & 3.33 & 5.70 & 8.02 \\
Qwen3-30B~\cite{yang2025qwen3} + PKG & 4.07 & 7.46 & 9.63 \\
PKG beam search & \underline{15.19} \scriptsize $\pm$ 1.70 & \underline{14.56} \scriptsize $\pm$ 2.46 & \underline{14.76} \scriptsize $\pm$ 2.46 \\
\midrule
PDPP~\cite{wang2023pdpp} & 13.11 \scriptsize $\pm$ 2.22 & 11.67 \scriptsize $\pm$ 1.75 & 11.23 \scriptsize $\pm$ 1.39 \\
KEPP~\cite{nagasinghe2024not} & 8.67 \scriptsize $\pm$ 1.93 & 8.77 \scriptsize $\pm$ 2.11 & 11.44 \scriptsize $\pm$ 1.93 \\
PlanLLM~\cite{yang2025planllm} & 9.70 \scriptsize $\pm$ 2.59 & 10.26 \scriptsize $\pm$ 1.05 & 11.55 \scriptsize $\pm$ 0.86 \\
SCHEMA~\cite{niu2024schema} & 10.37 \scriptsize $\pm$ 0.97 & 11.32 \scriptsize $\pm$ 2.19 & 12.19 \scriptsize $\pm$ 2.14 \\
\textbf{ViterbiPlanNet} & \textbf{19.04} \scriptsize $\pm$ 1.56 & \textbf{14.74} \scriptsize $\pm$ 1.75 & \textbf{16.90} \scriptsize $\pm$ 1.28 \\
\midrule
Improvement & +3.85 \scriptsize $\pm$ 2.37 & +0.18 \scriptsize $\pm$ 3.33 & +2.14 \scriptsize $\pm$ 2.67 \\
\bottomrule
\end{tabular}%
}
\end{table}

\subsection{Comparison with MTID ($fn \ 12$)}
Table~\ref{tab:performance_comparison_MTID_full} provides the extended comparison between ViterbiPlanNet$^{\clubsuit}$ and MTID$^{\clubsuit}$~\cite{zhou2025masked} across all reported horizons and datasets. Since MTID contains over one billion parameters and is computationally prohibitive to retrain, all MTID results are taken directly from the original paper (if available). To ensure fairness, we adapt ViterbiPlanNet to the MTID evaluation protocol, including the merged CrossTask taxonomy, modified mIoU computation, and PDPP-style feature extraction, and we denote these results with $^{\clubsuit}$. 

Across all datasets and horizons, the trends observed in the main paper remain consistent. Despite being \emph{three orders of magnitude smaller} (5--7M vs.\ 1,085M parameters), ViterbiPlanNet achieves performance comparable to MTID in terms of Success Rate and mean Accuracy, and substantially surpasses it in terms of mean IoU. At $T{=}3$ and $T{=}4$, ViterbiPlanNet delivers large improvements in mIoU, e.g., +7.75 points on CrossTask, +21.51 on COIN, and +38.61 on NIV at $T{=}3$, indicating that its predictions are considerably more locally accurate and structurally coherent. For longer horizons ($T{=}5$ and $T{=}6$), ViterbiPlanNet matches or exceeds the SR reported for MTID. 

These results confirm that ViterbiPlanNet achieves competitive or superior performance to MTID while being vastly more efficient, illustrating the benefits of integrating procedural structure directly into the decoding process rather than relying on large diffusion-based models.

\section{Additional Studies}

\subsection{Study on Rigidity of the Markov Assumption ($fn \ 1$)}
\begin{figure}[t]
    \centering
    \includegraphics[width=0.5\linewidth]{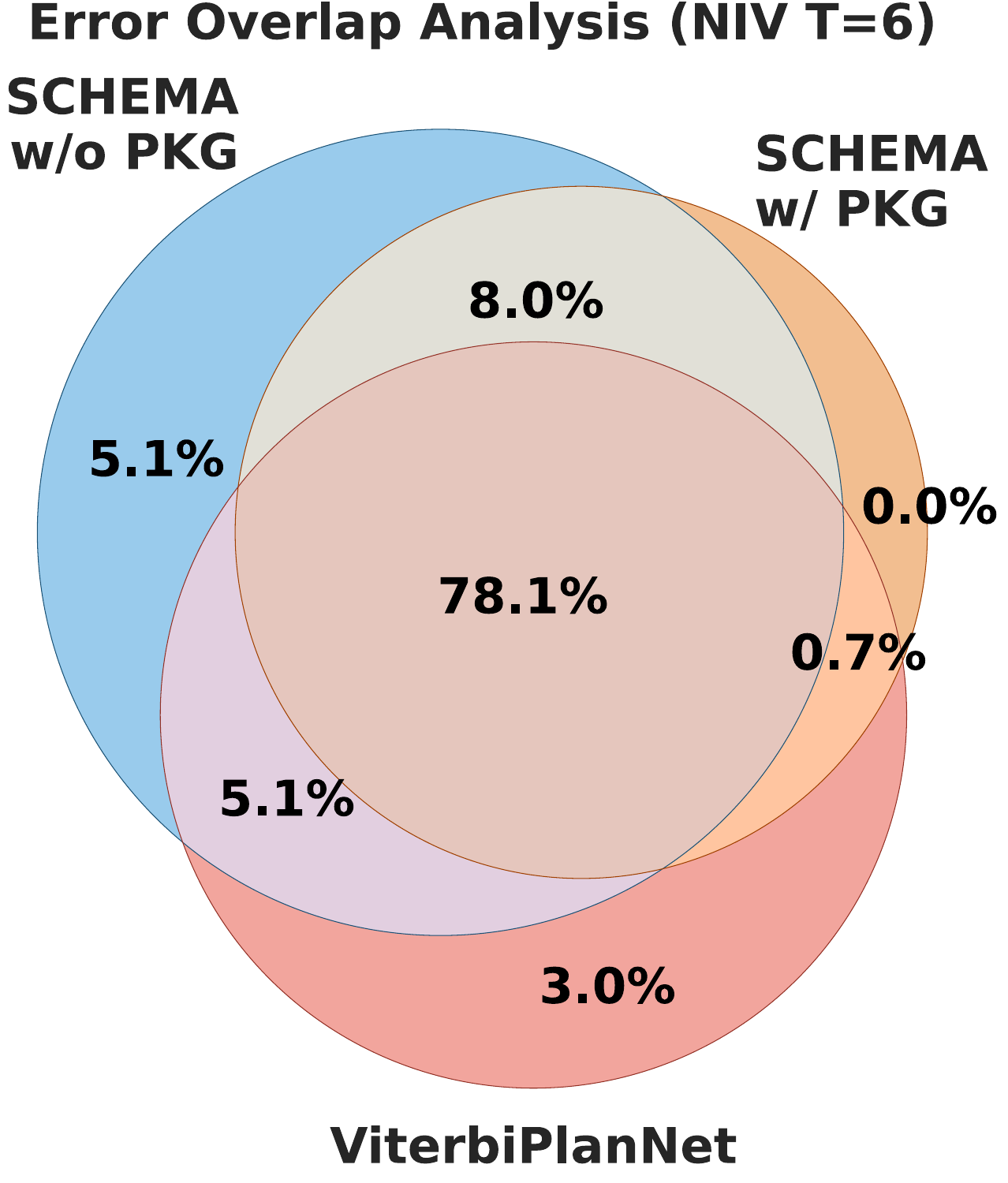}
    \caption{Error overlap analysis on NIV $T=6$.}
    \label{fig:venn_pkg}
\end{figure}

We investigate whether the Markov assumption enforced by Viterbi decoding introduces excessive rigidity during plan generation.
Our analysis evaluates three complementary aspects: (i) empirical coverage of the Procedural Knowledge Graph (PKG), 
(ii) decoding diversity under controlled conditions, and (iii) overlap of failure cases. 
Quantitative results are summarized in Table~\ref{tab:appendix_pkg_coverage} and Figure~\ref{fig:venn_pkg}.

\begin{table}[t]
\centering
\caption{PKG coverage and decoding diversity metrics.}
\label{tab:appendix_pkg_coverage}

    \resizebox{\linewidth}{!}{%
    \begin{tabular}{@{}lll@{}}
    \toprule
    \textbf{Metric} & \textbf{Dataset/Condition} & \textbf{Result} \\
    \midrule
    PKG Coverage & CrossTask ($T{=}3{\to}6$) & $97.4 \to 93.0$ (\%) \\
    (\% test sequences with & COIN ($T{=}3{\to}6$) & $90.8 \to 77.0$ (\%) \\
    transitions in the graph)& NIV ($T{=}3{\to}6$) & $86.7 \to 72.3$ (\%) \\
    \midrule
    Decoding Diversity & Uniform Emissions & $H{=}0.00$, JSD${=}0.56$ \\
    (Entropy $H{\uparrow}$, Jensen-- & SCHEMA w/o PKG & $H{=}4.39$, JSD${=}0.51$ \\
    Shannon Divergence JSD${\downarrow}$, & SCHEMA w/ PKG & $H{=}2.61$, JSD${=}0.39$ \\
    CrossTask, $T{=}6$) & ViterbiPlanNet & $H{=}2.13$, JSD${=}0.39$ \\
    \bottomrule
    \end{tabular}%
    }
\end{table}

\paragraph{PKG Coverage.}
We first measure PKG coverage, defined as the percentage of test trajectories whose transitions are present in the graph extracted from training data. 
Coverage naturally decreases with planning horizon and dataset complexity, ranging from $97.4\%\!\rightarrow\!93.0\%$ on CrossTask, 
$90.8\%\!\rightarrow\!77.0\%$ on COIN, and $86.7\%\!\rightarrow\!72.3\%$ on NIV for $T{=}3{\to}6$. 
Despite longer horizons, coverage remains consistently high ($>{}72\%$), indicating that Markovian decoding operates within a broadly permissive transition space. 
Uncovered transitions correspond primarily to procedures never observed during training, reflecting data sparsity rather than structural limitations of the decoder.

\paragraph{Decoding Diversity.}
To assess whether Viterbi decoding restricts procedural variability, we evaluate decoding diversity on a controlled subset of CrossTask ($T{=}6$), where all sequences share identical start and goal actions but differ in intermediate steps. 
We report entropy ($H{\uparrow}$) and Jensen--Shannon divergence from the ground-truth distribution ($\mathrm{JSD}{\downarrow}$).

A PKG-only decoder with uniform emissions collapses to a single trajectory ($H{=}0.00$), showing that transition constraints alone are insufficient to model procedural variability. 
Introducing learned emissions substantially increases diversity while maintaining alignment with ground truth. 
ViterbiPlanNet achieves $H{=}2.13$ and $\mathrm{JSD}{=}0.39$, comparable to SCHEMA with PKG ($H{=}2.61$, $\mathrm{JSD}{=}0.39$). 
Removing Viterbi decoding (SCHEMA w/o PKG) further increases entropy ($H{=}4.39$) but worsens distributional alignment ($\mathrm{JSD}{=}0.51$), indicating that additional trajectories are largely hallucinated rather than valid alternatives.

These results suggest that the Markov assumption acts as a \emph{structural prior}: it filters implausible transitions while learned emissions retain sufficient flexibility to explore multiple valid plans.

\paragraph{Failure Case Analysis.}
We further analyze error overlap on NIV at $T{=}6$ using the Venn diagram in Figure~\ref{fig:venn_pkg}. 
Adding PKG constraints to SCHEMA corrects $10.2\%$ ($5.1{+}5.1$) of failures while introducing only $0.7\%$ new errors, demonstrating that Markovian decoding primarily removes invalid trajectories rather than creating new failure modes. 
Despite architectural differences, ViterbiPlanNet introduces only $3\%$ new mistakes while correcting $13.1\%$ ($8{+}5.1$) and an additional $8\%$ of errors made by SCHEMA without and with PKG, respectively.

\paragraph{Discussion.}
Overall, the Markov assumption does not impose excessive rigidity. 
Instead, it provides a permissive yet structured decoding space enabled by high PKG coverage, within which learned emissions steer trajectory selection and recover correct procedural plans. 
This observation aligns with prior work (e.g., SCHEMA~\cite{niu2024schema}, PlanLLM~\cite{yang2025planllm}), where Viterbi decoding is routinely adopted as a final inference step.

\begin{figure}[t]
    \centering
    \includegraphics[width=\linewidth]{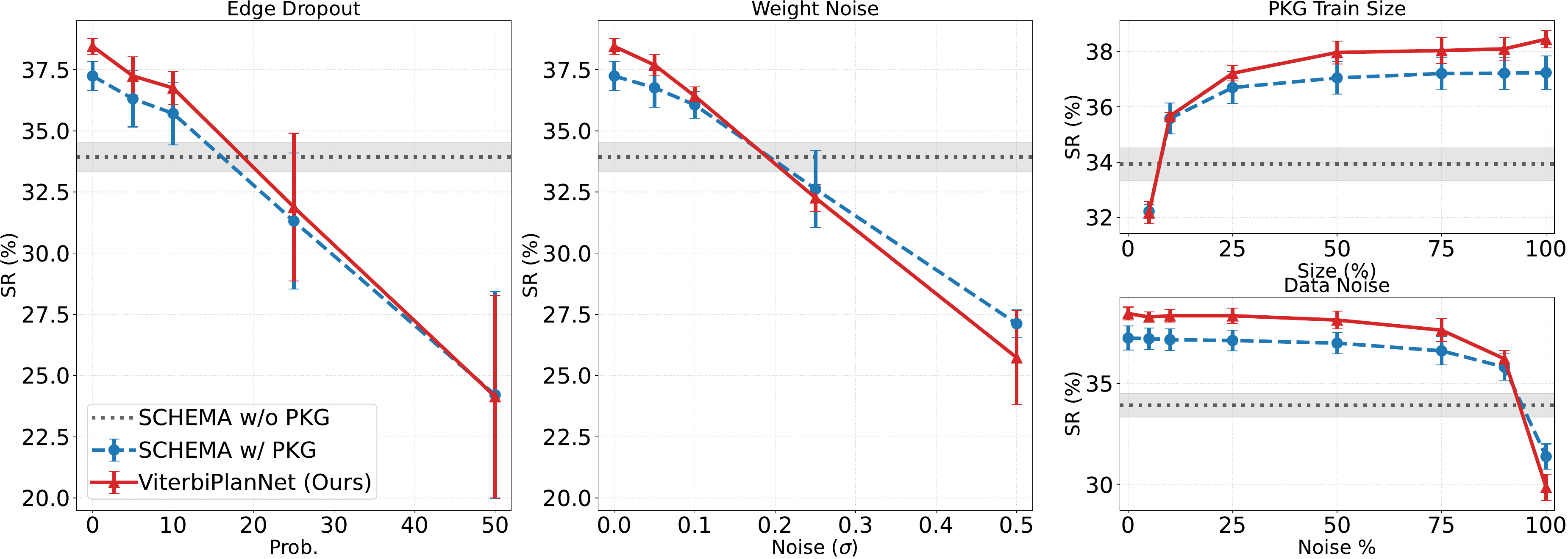}
    \caption{
    We evaluate performance under progressively corrupted procedural knowledge graphs (PKGs). 
    \textbf{Left:} edge dropout obtained by randomly removing graph transitions. 
    \textbf{Center:} noisy transition probabilities produced by Gaussian perturbations with variance $\sigma$. 
    \textbf{Right:} PKGs reconstructed from limited or noisy supervision.
    }
    \label{fig:noist_study}
\end{figure}

\subsection{Study on Dependence on PKG Quality ($fn \ 2$)}
We analyze the sensitivity of our method to the quality of the Procedural Knowledge Graph (PKG) by systematically degrading the graph used during training and inference. 
Specifically, we consider three complementary perturbations: 
(i) \emph{edge dropout}, obtained by randomly dropping transitions from the graph; 
(ii) \emph{edge-weight noise}, introduced by adding Gaussian perturbations to transition probabilities; and 
(iii) \emph{imperfect graph estimation}, where PKGs are reconstructed from limited or noisy supervision (Figure~\ref{fig:noist_study}).

Across all settings, ViterbiPlanNet exhibits graceful performance degradation. 
The model remains stable when removing up to $10$--$15\%$ of graph edges, injecting noise up to $\sigma=0.15$, or constructing PKGs from as little as $25\%$ of the available training data. 
Even under extreme label corruption (up to $90\%$ noisy annotations), performance remains competitive and consistently surpasses SCHEMA without PKG constraints (dashed curve).

Although the success rate decreases under severe corruption or extremely limited data, similar degradation trends are observed for SCHEMA with PKG decoding. 
This indicates that sensitivity primarily stems from inference with incomplete or noisy graphs rather than from our PKG-aware learning formulation.

\begin{table}[t]
    \caption{
    A single model is trained and evaluated using a unified PKG constructed from the union of all datasets.
    Gray values indicate the average performance of dataset-specific models, while arrows show the performance change when switching to a dataset-independent PKG.
    }
    \label{tab:dataset_independent}
    \resizebox{\linewidth}{!}{%
        \begin{tabular}{l r r r r}
        \toprule
        \textbf{Method}  
        [SR $\uparrow$ (\%)] 
        & $T=3$ & $T=4$ & $T=5$ & $T=6$\\
        \midrule
        SCHEMA~\cite{niu2024schema} 
        & \textcolor{gray}{32.1 $\to$} 32.7 \scriptsize$\pm$0.4 
        & \textcolor{gray}{23.6 $\to$} 21.9 \scriptsize$\pm$0.3 
        & \textcolor{gray}{16.6 $\to$} 14.2 \scriptsize$\pm$0.4 
        & \textcolor{gray}{13.1 $\to$} 10.2 \scriptsize$\pm$0.2 \\
        
        \textbf{ViterbiPlanNet} 
        & \textcolor{gray}{34.9 $\to$} \textbf{33.3} \scriptsize$\pm$0.4 
        & \textcolor{gray}{25.4 $\to$} \textbf{22.8} \scriptsize$\pm$0.2 
        & \textcolor{gray}{18.3 $\to$} \textbf{15.0} \scriptsize$\pm$0.2 
        & \textcolor{gray}{14.1 $\to$} \textbf{10.8} \scriptsize$\pm$0.2 \\
        \midrule
        
        Improvement 
        & +0.6 \scriptsize$\pm$0.5 
        & +0.8 \scriptsize$\pm$0.4 
        & +0.9 \scriptsize$\pm$0.5 
        & +0.6 \scriptsize$\pm$0.3 \\
        \bottomrule
        \end{tabular}%
    }
\end{table}

\subsection{Dataset-Independent PKG ($fn \ 3$)}
In Table~\ref{tab:dataset_independent} we evaluate whether procedural knowledge can generalize across datasets by training and testing a single model using a unified PKG constructed from the union of all datasets, instead of dataset-specific graphs.
Using a shared PKG introduces a modest performance drop compared to the average performance of models trained with dataset-specific graphs ($xx.x \rightarrow zz.z$ overall), reflecting the increased variability and partial mismatch between procedures originating from different domains. 
Nevertheless, ViterbiPlanNet consistently outperforms SCHEMA across all planning horizons, demonstrating that the learned emissions successfully adapt to heterogeneous procedural structures.

\subsection{Intermediate Visual Observations ($fn \ 4$)}
While planning in instructional videos is typically studied under a weakly supervised setting where only start and end observations are available~\cite{wang2023pdpp, niu2024schema, nagasinghe2024not, yang2025planllm}, real-world scenarios often provide additional intermediate visual cues that may refine or revise an ongoing plan. 
We therefore evaluate whether planning models can effectively incorporate partial observations appearing during execution.

To simulate plan revision, we train and evaluate both ViterbiPlanNet and SCHEMA on CrossTask with planning horizon $T{=}6$, providing as input the first three observed actions together with the final observation. 
This setting supplies partial trajectory evidence while leaving intermediate steps to be inferred, requiring the model to integrate new visual information with procedural priors.

Under this protocol, ViterbiPlanNet achieves a substantial improvement in Success Rate, increasing from $10.4 \pm 0.2$ to $18.3 \pm 0.7$. 
In comparison, SCHEMA improves from $10.2 \pm 0.4$ to $15.7 \pm 1.1$, resulting in a consistent performance gap in favor of our approach. 
The larger gain indicates that ViterbiPlanNet uses additional visual context more effectively to refine trajectory predictions.

\begin{table}[t]
    \caption{
    Evaluation on the egocentric instructional video dataset EgoPER~\cite{lee2024error} across planning horizons $T{=}3$--$6$.
    }
    \label{tab:egoper_results}
    \resizebox{\linewidth}{!}{%
    \begin{tabular}{l l l l l}
    \toprule
    \textbf{Method}  
    [SR $\uparrow$ (\%)]
    & $T=3$ & $T=4$ & $T=5$ & $T=6$ \\
    \midrule
    SCHEMA~\cite{niu2024schema} 
    & 29.55 \scriptsize$\pm$1.95 
    & 20.78 \scriptsize$\pm$1.94 
    & 13.91 \scriptsize$\pm$1.10 
    & 12.46 \scriptsize$\pm$1.57 \\
    \textbf{ViterbiPlanNet} 
    & \textbf{51.84} \scriptsize$\pm$0.95 
    & \textbf{48.14} \scriptsize$\pm$0.64 
    & \textbf{46.34} \scriptsize$\pm$0.49 
    & \textbf{41.98} \scriptsize$\pm$0.92 \\
    \midrule
    Improvement 
    & +22.29 \scriptsize$\pm$2.17 
    & +27.36 \scriptsize$\pm$2.04 
    & +32.43 \scriptsize$\pm$1.21 
    & +29.52 \scriptsize$\pm$1.82 \\
    \bottomrule
    \end{tabular}%
    }
\end{table}

\subsection{Planning in Egocentric Instructional Videos ($fn \ 7$)}

In Table~\ref{tab:egoper_results} we further evaluate our approach on \textbf{EgoPER}~\cite{lee2024error}, a challenging egocentric instructional video dataset characterized by long-horizon tasks, strong viewpoint changes, and increased visual ambiguity compared to third-person instructional benchmarks. 
In this setting, action observations are captured from a first-person perspective, making procedural reasoning more difficult due to partial visibility, motion-induced noise, and higher intra-class variability.
ViterbiPlanNet achieves substantial improvements over SCHEMA across all planning horizons ($T{=}3$--$6$). 
Performance gains range from $+22.3\%$ to $+32.4\%$ absolute Success Rate, indicating that structured Markov decoding combined with learned visual emissions remains effective even under severe viewpoint variability.

These results demonstrate that our method generalizes beyond third-person instructional videos and transfers effectively to egocentric planning scenarios. 
The findings support the hypothesis that procedural structure provides a domain-agnostic inductive bias, enabling robust planning despite substantial visual distribution shifts.

Extending evaluation to robotics and interactive environments with stochastic or branching procedures represents an important direction for future work.

\section{Further Experimental Details}

\subsection{Hyperparameter Configuration}

\paragraph{Baseline Configurations.}
All baselines are implemented following the configurations reported in their original papers, with one exception: PlanLLM~\cite{yang2025planllm}. Upon inspection, we found that the released implementation applies a self-attention module over the start, end, and intermediate frames during both training and inference. Since intermediate frames cannot be used at test time, this introduces an unintended information leak. We therefore corrected this behavior by ensuring that PlanLLM attends only to the start and end frames at inference, aligning it with the standard protocol~\cite{bi2021procedure} and ensuring a fair comparison.

\begin{table} [t]
    \caption{ViterbiPlanNet training configuration.}
    \label{tab:hyperparameters}
    \centering
    \begin{tabular}{lc}
    \toprule
         Component & Name/Value \\
    \midrule
         Visual Backbone & S3D~\cite{xie2018rethinking} \\
         Optimizer & Adam~\cite{adam2014method} \\
         Learning Rate & $9 \times 10^{-3}$ \\
         Dropout & 0.20 \\
         Batch Size & 256 \\
         Epochs & 500 \\
         Embedding Dimension ($E$) & 128 \\
         
    \bottomrule
    \end{tabular}
\end{table}

\paragraph{ViterbiPlanNet Configuration.} Table~\ref{tab:hyperparameters} reports the complete set of hyperparameters used to train ViterbiPlanNet across all datasets. Unless otherwise specified, the same configuration is applied across all planning horizons. Visual representations are extracted using the S3D backbone~\cite{xie2018rethinking}, followed by a projection layer, a lightweight Transformer encoder, an MLP with a Sigmoid activation, and finally the Differentiable Viterbi Layer (DVL).
Training is performed with the Adam optimizer~\cite{adam2014method}, using a learning rate of $9 \times 10^{-3}$, a dropout rate of $0.20$, and a batch size of $256$. Models are trained for $500$ epochs, which we found sufficient for stable and consistent convergence across datasets and horizons. The embedding dimensionality for action representations is fixed to $E = 128$. 
The only deviation from this configuration occurs in the experiments reported in Table~\ref{tab:horizon_comparison_coin}: for COIN at $T{=}6$, we employ two concatenated DVL layers during training. This additional depth improves normalization stability and enhances cross-horizon consistency on this particularly challenging dataset.

\subsection{LLM and VLM Details ($fn \ 12$)}
\begin{figure*}[t]
    \centering
    \includegraphics[width=\linewidth]{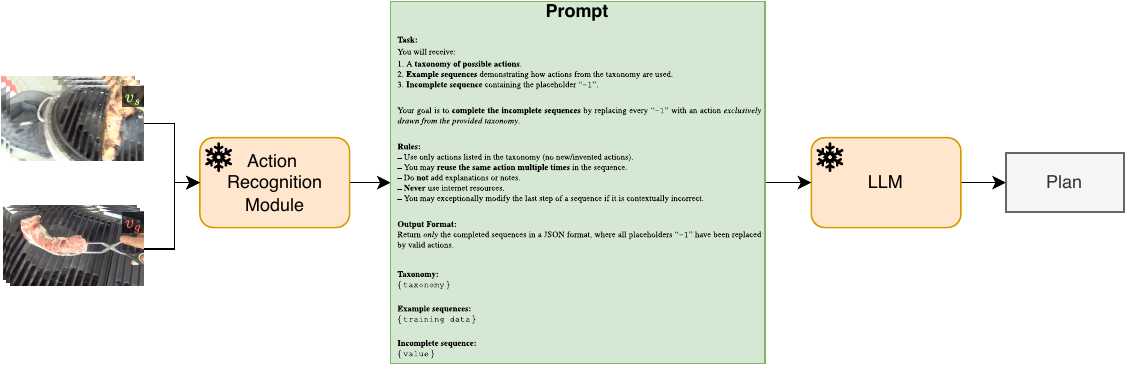}
    \caption{LLM-Based sequence completion approach.}
    \label{fig:prompt_LLM}
\end{figure*}

\begin{figure*}[t]
    \centering
    \includegraphics[width=\linewidth]{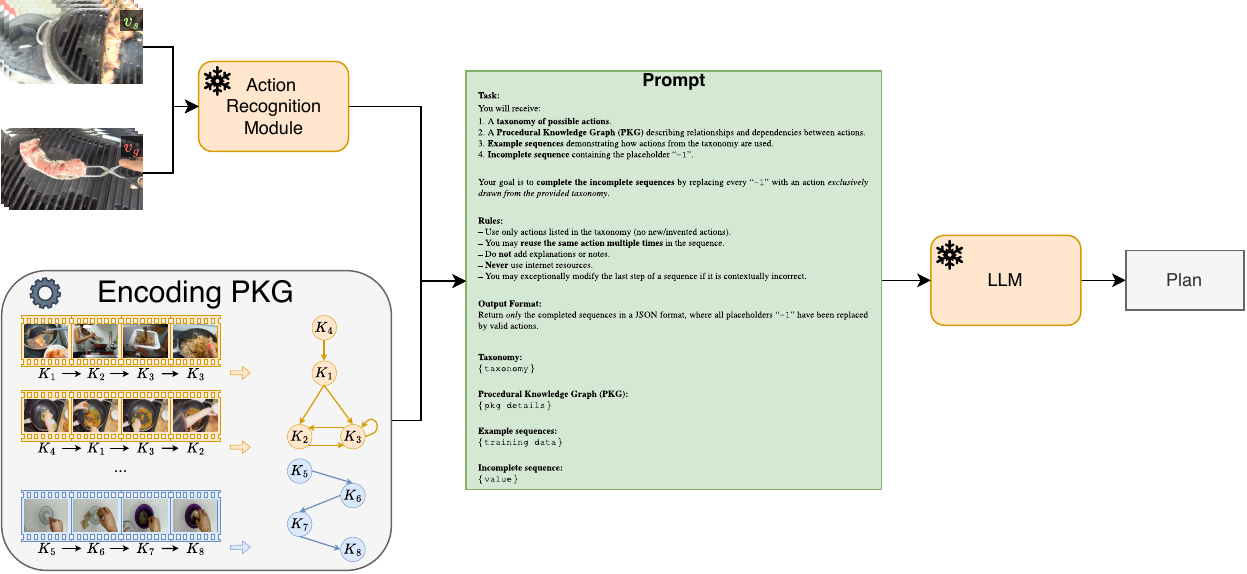}
    \caption{LLM + PKG sequence completion approach.}
    \label{fig:prompt_LLM_PKG}
\end{figure*}

\begin{figure*}[t]
    \centering
    \includegraphics[width=\linewidth]{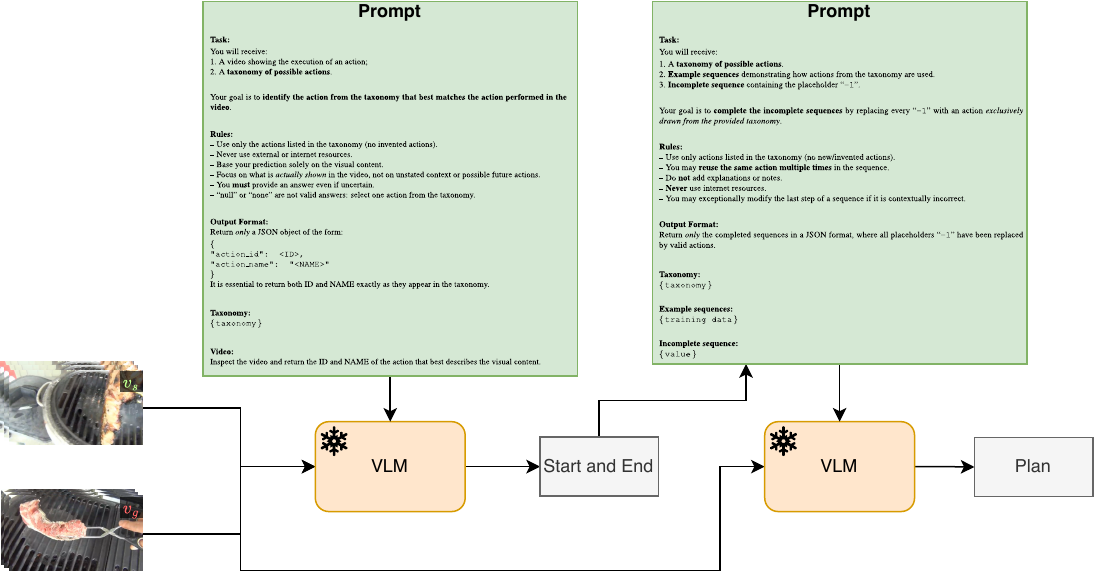}
    \caption{VLM-based sequence completion approach.}
    \label{fig:prompt_VLM}
\end{figure*}

This section details how Large Language Models (LLMs), and Vision-Language Models (VLMs) are employed for planning in instructional videos.  
Each family of models follows a dedicated strategy, illustrated in Figures~\ref{fig:prompt_LLM}, \ref{fig:prompt_LLM_PKG}, and \ref{fig:prompt_VLM}.

\paragraph{LLM-Based Sequence Completion.} LLMs are used to complete partially observed action sequences given an action taxonomy and several example trajectories.  
As shown in Figure~\ref{fig:prompt_LLM}, the model receives: (1) a taxonomy of admissible actions, (2) example sequences demonstrating valid procedural structure, and (3) an incomplete sequence containing the placeholder ``\texttt{-1}''.  

The LLM is instructed to substitute each missing element using only actions from the taxonomy, without generating new actions or explanations. The model may reuse actions multiple times and may adjust the final step if contextually inconsistent.

\paragraph{LLM + PKG Sequence Completion.} When a Procedural Knowledge Graph (PKG) is available, we further constrain the LLM with structural priors (Figure~\ref{fig:prompt_LLM_PKG}).  
In addition to the taxonomy and training examples, the model receives the PKG, which encodes known action dependencies and admissible transitions following~\cite{fatemi2024talk}.

\paragraph{VLM Sequence Completion.}
For Vision--Language Models, we adopt a two-stage approach (Figure~\ref{fig:prompt_VLM}). 

In the first stage, the VLM is provided with:
(1) the action taxonomy, and 
(2) a video clip depicting the execution of the \emph{start} and \emph{end} actions. 
The model must identify and return exactly two actions from the taxonomy.

In the second stage, the predicted start and end actions are inserted into Prompt~\ref{prompt:sequence_completion}, and the VLM is then queried again, now conditioned on the video and the updated prompt, to generate the full action sequence. 
This two-step strategy enables the VLM to produce a complete plan while remaining grounded in the visual evidence and constrained by the predefined action taxonomy.

\paragraph{Prompts.} Prompts~\ref{prompt:sequence_completion}, \ref{prompt:sequence_completion_pkg}, and \ref{prompt:vlm_action} reproduce the exact prompts used for LLMs, LLMs+PKG, and VLMs, respectively, ensuring reproducibility of our experimental setup.

\newpage

\begin{prompt*}
\centering
\fbox{
    \begin{minipage}{0.85\linewidth}
    \textbf{Task:}\\[2pt]
    You will receive:\\[2pt]
    1.\;A \textbf{taxonomy of possible actions}.\\
    2.\;\textbf{Example sequences} demonstrating how actions from the taxonomy are used.\\
    3.\;\textbf{Incomplete sequence} containing the placeholder ``\texttt{-1}''.\\[6pt]

    Your goal is to \textbf{complete the incomplete sequences} by replacing every ``\texttt{-1}'' with an action \emph{exclusively drawn from the provided taxonomy}.\\[6pt]

    \textbf{Rules:}\\
    -- Use only actions listed in the taxonomy (no new/invented actions).\\
    -- You may \textbf{reuse the same action multiple times} in the sequence.\\
    -- Do \textbf{not} add explanations or notes.\\
    -- \textbf{Never} use internet resources.\\
    -- You may exceptionally modify the last step of a sequence if it is contextually incorrect.\\[6pt]

    \textbf{Output Format:}\\
    Return \emph{only} the completed sequences in a JSON format, where all placeholders
    ``\texttt{-1}'' have been replaced by valid actions.\\[10pt]

    \textbf{Taxonomy:}\\
    \texttt{\{\,taxonomy\,\}}\\[4pt]

    \textbf{Example sequences:}\\
    \texttt{\{\,training data\,\}}\\[4pt]

    \textbf{Incomplete sequence:}\\
    \texttt{\{\,value\,\}}
    \end{minipage}
}
\caption{Prompt used to instruct the model to complete action sequences from a predefined taxonomy.}
\label{prompt:sequence_completion}
\end{prompt*}

\begin{prompt*}
\centering
\fbox{
    \begin{minipage}{0.85\linewidth}
    \textbf{Task:}\\[2pt]
    You will receive:\\[2pt]
    1.\;A \textbf{taxonomy of possible actions}.\\
    2.\;A \textbf{Procedural Knowledge Graph (PKG)} describing relationships and dependencies between actions.\\
    3.\;\textbf{Example sequences} demonstrating how actions from the taxonomy are used.\\
    4.\;\textbf{Incomplete sequence} containing the placeholder ``\texttt{-1}''.\\[6pt]

    Your goal is to \textbf{complete the incomplete sequences} by replacing every ``\texttt{-1}'' with an action \emph{exclusively drawn from the provided taxonomy}.\\[6pt]

    \textbf{Rules:}\\
    -- Use only actions listed in the taxonomy (no new/invented actions).\\
    -- You may \textbf{reuse the same action multiple times} in the sequence.\\
    -- Do \textbf{not} add explanations or notes.\\
    -- \textbf{Never} use internet resources.\\
    -- You may exceptionally modify the last step of a sequence if it is contextually incorrect.\\[6pt]

    \textbf{Output Format:}\\
    Return \emph{only} the completed sequences in a JSON format, where all placeholders
    ``\texttt{-1}'' have been replaced by valid actions.\\[10pt]

    \textbf{Taxonomy:}\\
    \texttt{\{\,taxonomy\,\}}\\[4pt]

    \textbf{Procedural Knowledge Graph (PKG):}\\
    \texttt{\{\,pkg details\,\}}\\[4pt]

    \textbf{Example sequences:}\\
    \texttt{\{\,training data\,\}}\\[4pt]

    \textbf{Incomplete sequence:}\\
    \texttt{\{\,value\,\}}
    \end{minipage}
}
\caption{Prompt used when procedural knowledge graph (PKG) information is available for sequence completion.}
\label{prompt:sequence_completion_pkg}
\end{prompt*}

\begin{prompt*}
\centering
\fbox{
    \begin{minipage}{0.85\linewidth}
    \textbf{Task:}\\[2pt]
    You will receive:\\
    1.\;A video showing the execution of an action;\\
    2.\;A \textbf{taxonomy of possible actions}.\\[6pt]

    Your goal is to \textbf{identify the action from the taxonomy that best matches the action performed in the video}.\\[8pt]

    \textbf{Rules:}\\
    -- Use only the actions listed in the taxonomy (no invented actions).\\
    -- Never use external or internet resources.\\
    -- Base your prediction solely on the visual content.\\
    -- Focus on what is \emph{actually shown} in the video, not on unstated context or possible future actions.\\
    -- You \textbf{must} provide an answer even if uncertain.\\
    -- ``null'' or ``none'' are not valid answers: select one action from the taxonomy.\\[8pt]

    \textbf{Output Format:}\\
    Return \emph{only} a JSON object of the form:\\[2pt]
    \texttt{\{}\\
    \quad\texttt{"action\_id": <ID>,}\\
    \quad\texttt{"action\_name": "<NAME>"}\\
    \texttt{\}}\\
    It is essential to return both ID and NAME exactly as they appear in the taxonomy.\\[10pt]

    \textbf{Taxonomy:}\\
    \texttt{\{\,taxonomy\,\}}\\[10pt]

    \textbf{Video:}\\
    Inspect the video and return the ID and NAME of the action that best describes the visual content.
    \end{minipage}
}
\caption{Prompt used for VLM-based action identification from video given a fixed action taxonomy.}
\label{prompt:vlm_action}
\end{prompt*}

\end{document}